\documentclass[11pt, a4paper, copyright, goog]{google}

\usepackage{times}
\usepackage{natbib}
\usepackage[utf8]{inputenc}
\usepackage[T1]{fontenc}
\usepackage{hyperref}
\usepackage{url}
\usepackage{booktabs}
\usepackage{amsfonts}
\usepackage{amsmath,amssymb}
\usepackage{bbm}
\usepackage{nicefrac}
\usepackage{microtype}
\usepackage{xcolor}
\usepackage{enumitem}
\usepackage{graphicx}
\usepackage{tikz}
\usepackage{pgfplots}
\pgfplotsset{compat=1.17}
\usepackage{tabularx}
\usepackage{makecell}
\usepackage{multirow}
\usepackage{algorithm}
\usepackage{algorithmic}
\usepackage{subcaption}
\usepackage{pifont}
\newcommand{\cmark}{\ding{51}}
\newcommand{\xmark}{\ding{55}}
\usetikzlibrary{arrows.meta,positioning,shapes,fit,calc}

\usepackage{xspace}
\newcommand{\methodname}{\textsc{Harmony Harness}\xspace}
\newcolumntype{L}{>{\raggedright\arraybackslash}X}

\newcommand{\todo}[1]{{\color{red}\textbf{[TODO: #1]}}}

\definecolor{fastcolor}{RGB}{0,128,0}
\definecolor{slowcolor}{RGB}{192,0,0}

\tikzset{
  block/.style={rectangle, draw, rounded corners, align=center,
                minimum width=3cm, minimum height=0.9cm},
  line/.style={draw, -{Latex}},
}

\hypersetup{
  colorlinks=true,
  linkcolor=blue!60!black,
  citecolor=blue!60!black,
  urlcolor=blue!60!black,
}

\title{Harmony Harness: Multi-Objective Self-Evolution of Agent Harnesses}

\correspondingauthor{wyu3@andrew.cmu.edu, lmiculicich@google.com, and longtle@google.com}

\author[1 2 *]{Weichen Yu}
\author[1]{Lesly Miculicich}
\author[1]{Bhavana Dalvi Mishra}
\author[1]{Jiefeng Chen}
\author[3]{Shuo Li}
\author[2]{Matt Fredrikson}
\author[1]{Tomas Pfister}
\author[1]{Long T. Le}

\affil[1]{Google Cloud AI Research}
\affil[2]{Carnegie Mellon University}
\affil[3]{Google DeepMind}

\begin{abstract}
Recent work on self-evolving LLM agents treats the agent harness as an optimization target that can be iteratively improved through execution feedback. Existing approaches, however, largely optimize a single objective and evolve the harness as a monolithic artifact, limiting their ability to handle competing goals and to separate transferable design from environment-specific adaptation. We introduce \methodname, a framework for \emph{multi-objective, compositional harness evolution}. \methodname formulates harness evolution as a multi-objective optimization problem and uses decomposition-based evolutionary search to maintain solutions across different regions of the Pareto frontier, avoiding reliance on a hand-designed structural diversity metric for repository-scale harnesses. In parallel, we introduce factorized evolution operators that separately refine general-purpose harness mechanisms and benchmark-specific adaptations, together with a dynamic composition workflow that integrates them at inference time. Across SecRepoBench and BaxBench, \methodname consistently outperforms strong harness-evolution baselines, with gains exceeding 10\% on both benchmarks. Our results show that explicitly modeling both objective trade-offs and harness structure yields substantially stronger and more transferable agent evolution than existing single-objective, monolithic approaches.
\end{abstract}

\begin{document}

\maketitle
\section{Introduction}

A deployed LLM agent wraps a base model in a \emph{harness}: the system
prompt, tool set, agent loop, verification gates, and orchestration
that collectively determine how underlying capabilities translate into task outcomes.
Recent work on \emph{self-evolving harnesses} allows agents to iteratively modify
this infrastructure from execution feedback~\citep{hu2024adas,zhuge2024gptswarm,zhang2026darwin,robeyns2025sica,zhang2026selfharness,qiu2025alita,lee2026meta}.
Existing approaches, however, optimize solely around a \emph{single scalar objective},
such as pass rate on a benchmark. \textbf{Harness self-evolution under
\emph{multiple} objectives, in particular with \emph{safety} treated as an
independent axis rather than folded into the success signal, remains unaddressed.}

Extending harness self-evolution from a single objective to multiple objectives is not as simple as adding another reward signal. The challenge arises from \textbf{the structure of the Pareto frontier and the possibility of genuine trade-offs between objectives}. Consider functionality and safety. At first glance, the two appear aligned: a stronger coding agent should ideally produce code that both passes its tests and resists exploitation. Yet this alignment need not hold throughout the space of possible harnesses. A harness may improve pass rate by relaxing a verification gate, for example, or improve security by rejecting behaviors that are nevertheless required for functional correctness. Once such conflicts emerge, the evolution procedure must explicitly determine how they should be resolved. Existing formulations handle such tensions poorly:
linear scalarization provably misses non-convex Pareto regions\footnote{The Pareto frontier is the set of non-dominated solutions for which no objective can be improved without degrading at least one other objective.} regardless of weight choices, and, even where the frontier is convex, it will trade a point of one objective for a cheaper point of the other;
delegating trade-offs to the agent leaves compromises implicit, unstable,
and rarely Pareto-optimal. Constrained optimization~\citep{sinha2024survival,dai2023safe} enforces
safety thresholds, but reduces the task to constrained single-objective search rather
than mapping the trade-off surface. Handling multiple objectives thus requires an
explicit mechanism that keeps pressure on \emph{both} axes. Whether the frontier
is actually non-convex is an empirical question: on the secure-coding benchmarks
we study, better scaffolds often improve functionality and security together
(App.~\ref{sec:appendix-tables}). There, the main value of
decomposition is not recovering non-convex regions. It is objective-directed search
pressure: each subproblem keeps pushing whichever objective has the largest
weighted shortfall from the ideal point, so neither objective can regress unnoticed.

A second limitation of existing harness-evolution methods is structural rather than algorithmic: they evolve the harness as a \emph{monolithic} artifact. Each generation rewrites the repository as a single block against feedback from a single environment; 
\textbf{they do not disentangle general harness design from environment-specific adaptation} \citep{lee2026meta,yin2025godel}. In practice, many mechanisms acquired during evolution contain both components. One is a \emph{general capability} that should transfer across environments; the other is the \emph{environment-specific calibration} required to make that capability effective in a particular setting. For example, constructing and executing code inside a container is a broadly useful capability, whereas the choice of image, teardown policy, and handling of benchmark-specific sandbox conventions is environment dependent. Similarly, compressing or truncating tool outputs is a general mechanism, while deciding which portions can safely be omitted depends strongly on whether the output is a compiler log, a fuzzer trace, or a test-runner report. Current methods typically optimize these components jointly against feedback from a single environment. As a result, transferable design choices and benchmark-specific adaptations receive the same reward signal and are encoded into the same monolithic harness. We do not expect the boundary between these two designs to be perfectly discrete; effective capabilities inevitably require some environment-dependent calibration. Nevertheless, explicitly separating the two creates a more useful abstraction: a \emph{general harness} that can transfer across tasks and environments, paired with lightweight \emph{environment-specific components} that specialize it to the execution context at hand.

To address these challenges, we introduce \methodname, a \emph{multi-objective harness evolution} framework built on decomposition-based evolutionary optimization~\citep{li2009moeadde}. Rather than collapsing multiple objectives into a single reward, we decompose the objective space using a set of weight vectors, each defining a scalar subproblem corresponding to a subregion of the Pareto frontier, and optimize these subproblems jointly within a shared population. This formulation is particularly well suited to harness evolution. Existing evolutionary systems often require an explicit artifact-level diversity metric to prevent population collapse~\citep{novikov2025alphaevolve,wang2025thetaevolve}; such metrics (e.g. number of lines of code) are relatively natural for compact artifacts such as a single program, but become difficult to define meaningfully for repository-scale agent harnesses whose variations span prompts, control flow, tools, verification logic, and orchestration. Our decomposition instead gives each subproblem a different objective preference, so selection pressure is diversified by construction, without requiring a hand-designed structural diversity metric. This does not by itself guarantee \emph{behavioral} diversity: when the objectives are positively correlated, or when LLM edits are small and local, the incumbents need not differ in behavior. We find that standard code-level diversity metrics capture behavioral diversity only partially for harnesses (appendix). Maintaining a bounded set of representative solutions also avoids carrying an ever-growing archive of artifacts into the proposer context, reducing the context and search overhead of evolution.

We instantiate dual-objective self-evolution in \emph{secure code generation}, where functionality
requires passing unit tests and security requires neutralizing proof-of-concept exploits.
Our findings are: 1) \textbf{Limits of Single-Objective Evolution:} Optimizing solely for code functionality yields only marginal, incidental security gains. 2) \textbf{Suboptimality of Naive Formulations:} Naive combinations, such as weighted-sum scalarization or sequential evolution, stall below the full framework; the margin comes from combining decomposition with factored evolving and dynamic orchestration, not from the aggregation rule alone.
3) \textbf{Effectiveness and Transferability:} Our multi-objective framework consistently outperforms baselines and generalizes across benchmarks with identical dual objectives.
4) \textbf{Benefits of Factorized Harness and Dynamic Orchestration:} Factorized harness evolution and adaptive orchestration outperform static, fixed-pipeline evolutionary workflows.

\section{Related Work}

\textbf{Self-evolving and self-improving agents.}
A growing body of work lets agents improve from their own experience via
reflection, memory, skill libraries, and workflow
updates~\citep{wang2023voyager,shinn2023reflexion,sun2023adaplanner,tang2024agentpro,ou2025symbolic,zhang2025aflow,hu2024evomac}.
Voyager~\citep{wang2023voyager} grows a skill library from iterative code;
Reflexion~\citep{shinn2023reflexion} uses verbal self-critique over failed
trajectories; AdaPlanner~\citep{sun2023adaplanner} adapts plans from
environmental feedback. Two systems are conceptual predecessors of ours; we
borrow ideas from them, not their algorithms or code.
GEPA~\citep{agrawal2026gepa} performs reflective prompt evolution over a
maintained Pareto frontier of prompts; we lift the idea of a maintained frontier
from prompts to whole harnesses and from task instances to a
$(\mathrm{functionality},\mathrm{security})$ objective space. ADAS~\citep{hu2024adas}
has a meta-agent program new agent designs in code, one level up from the
prompt; we share its view of the agent as an editable program, but our search
procedure is MOEA/D~\citep{li2009moeadde} with LLM variation operators
(\S\ref{sec:moead}), not ADAS's archive-based meta-agent search. Closest to our code setting,
the Darwin G\"odel Machine~\citep{zhang2026darwin} and
meta-harness~\citep{lee2026meta} self-improve
a coding agent by a \emph{monolithic} loop---evolve the
agent's code as one block. All of these systems (i) optimize a
\emph{single} objective and evaluate only functionality, and (ii) do not separate
reusable from environment-specific components in \emph{what} is evolved.

\textbf{Safety during evolution, and safety as a constraint.}
SoS~\citep{sinha2024survival} jointly evolves a prompt for utility and safety, but
evolves only a single prompt slot; safety gains at one position can be undone
elsewhere. More broadly, multi-objective alignment treats
helpfulness/harmlessness trade-offs as joint reward
optimization~\citep{askell2021hhh,bai2022rlhf,bai2022constitutional,rame2023rewardedsoups,zhou2024beyond},
whereas constrained approaches argue safety is better expressed as a hard
constraint than a reward term~\citep{achiam2017cpo,dai2023safe}. Rather than demoting safety to a feasibility condition on a
functionality objective, we keep it as an objective in its own right and search
the frontier by decomposition, lifted from a single prompt to an entire harness.

\section{Problem Formulation}
\label{sec:formulation}

Let $\mathcal{H}$ denote the space of agent harnesses, and let
$\mathcal{D}=\{t_n=(x_n,p_n)\}_{n=1}^{M}$ be a set of $M$ evaluation
tasks, where $x_n$ is the task specification and $p_n$ is the corresponding
security policy. Executing harness $H\in\mathcal{H}$ on task $t_n$ induces a
distribution over trajectories,
$\tau \sim P_H(\tau \mid x_n),$
reflecting the stochasticity of the underlying agent. The agent only sees
$x_n$; the policy $p_n$ (e.g., the exploit) is hidden at solve time and enters
only through the evaluation oracle $S(\tau,p_n)$.
We evaluate each harness along two execution-grounded objectives,
functionality and security:
$
    \mathbf{f}(H)
    =
    \bigl(
        f_{\mathrm{func}}(H),
        f_{\mathrm{sec}}(H)
    \bigr),
$ defined in \ref{append:formulations}.

\textbf{Joint metric.} Our headline metric is the rate at which
a single trajectory is \emph{both} functional and secure:
\begin{equation}
    f_{\wedge}(H)
    =
    \frac{1}{M}\sum_{n=1}^{M}
    \mathbb{E}_{\tau\sim P_H(\cdot\mid x_n)}
    \left[F(\tau,x_n)\,S(\tau,p_n)\right].
    \label{eq:joint}
\end{equation}
The search optimizes the two marginals $(f_{\mathrm{func}},f_{\mathrm{sec}})$
rather than $f_{\wedge}$ directly. The joint rate is bounded on both sides by the
marginals (the Fr\'echet bounds),
\begin{equation}
    \max\bigl(0,\;f_{\mathrm{func}}(H)+f_{\mathrm{sec}}(H)-1\bigr)
    \;\le\;
    f_{\wedge}(H)
    \;\le\;
    \min\bigl(f_{\mathrm{func}}(H),f_{\mathrm{sec}}(H)\bigr),
    \label{eq:frechet}
\end{equation}
so raising both marginals raises the ceiling on $f_{\wedge}$ and, once their
sum exceeds $1$, also forces it up from below. A Tchebycheff subproblem
(Eq.~\ref{eq:tchebycheff}, \S\ref{sec:moead}) acts on the objective with the
largest \emph{weighted} shortfall from the ideal point; for the balanced weight
$\boldsymbol{\lambda}=(\tfrac12,\tfrac12)$ this is the currently worse
objective, which is the one that caps $f_{\wedge}$ through the upper bound. Optimizing the marginals
does not guarantee joint success: two harnesses can have the same marginals and
very different $f_{\wedge}$ if they succeed on different tasks. We therefore
report $f_{\wedge}$ and the gap $\min(f_{\mathrm{func}},f_{\mathrm{sec}})-f_{\wedge}$
for every method (\S\ref{sec:exp-strategy}). We keep the marginals as the search
objectives because they tell the proposer \emph{which} property failed on a task,
whereas $f_{\wedge}$ alone does not.

\textbf{Selecting the deployed harness.} Comparisons against single-harness
baselines, held-out evaluation, and transfer need one harness per run. Let
$\mathcal{U}_g$ be the set of children evaluated in generation $g$, and let
$\mathcal{U}=\bigcup_g \mathcal{U}_g$. After evolution we select the
\emph{most balanced} child of the run,
\begin{equation}
    H^{\dagger}
    =
    \arg\max_{H\in\mathcal{U}}
    \min\bigl(\hat f_{\mathrm{func}}(H),\hat f_{\mathrm{sec}}(H)\bigr),
    \label{eq:selection}
\end{equation}
computed on the search split only, with ties resolved in favor of the earlier
generation. The candidate need not be an incumbent at the end of the run: a child
that was later replaced under one subproblem's preference can still be the most
balanced one overall. The held-out splits are never used for selection.
Eq.~\ref{eq:selection} is the Tchebycheff scalarization
(Eq.~\ref{eq:tchebycheff}, \S\ref{sec:moead}) with equal weights
$\boldsymbol{\lambda}=(\tfrac12,\tfrac12)$ and the utopia point $(1,1)$ as
reference, since
$\arg\min_{H}\max_j\tfrac12\bigl(1-\hat f_j(H)\bigr)
=\arg\max_{H}\min_j\hat f_j(H)$.

\section{Multi-Objective and Factored Harness Evolution}
\label{sec:method}

\begin{figure}[t]
\centering
\includegraphics[width=\linewidth]{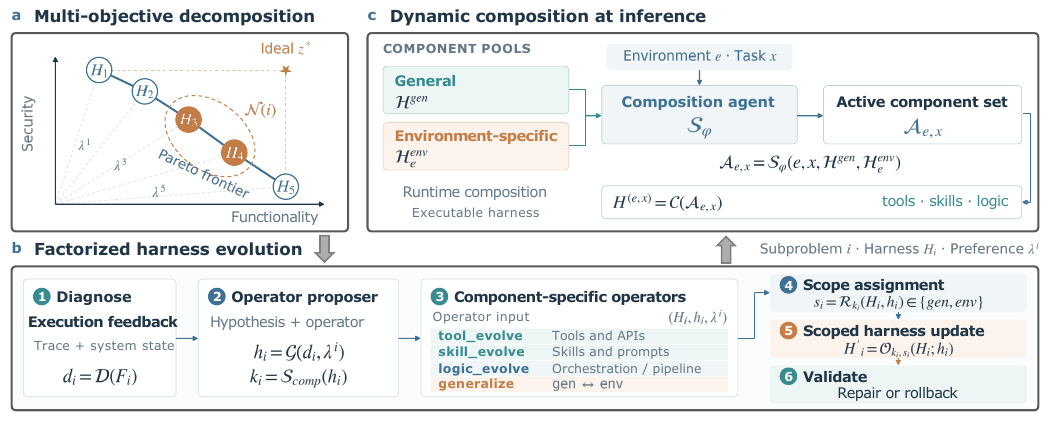}
\vspace{-1em}
\caption{\textbf{Overview of \textsc{Harmony Harness}.}
\textbf{(a)} The functionality--security space is split into weighted subproblems. Each subproblem keeps a harness and together they approximate the Pareto frontier.
\textbf{(b)} An LLM proposer diagnoses failures and forms a hypothesis. Operators edit tools, skills, or logic, and place each edit in either the general scope or the environment-specific scope. The child is validated before evaluation.
\textbf{(c)} At inference, a composition agent picks components from the general and environment-specific pools to build a task-specific harness.}
\label{fig:method-overview}
\vspace{-0.5cm}
\end{figure}

Given the multi-objective formulation in \S\ref{sec:formulation},
\methodname evolves a population of repository-scale agent harnesses toward
different regions of the Pareto frontier. The framework combines two
complementary ideas. First, we adapt decomposition-based multi-objective
evolution to LLM agent harnesses: a set of objective preferences defines
multiple scalar subproblems, each maintaining an incumbent harness for a
different trade-off between objectives. An LLM proposer generates candidate
harnesses from execution-grounded feedback, and candidates selectively replace
incumbents in nearby subproblems according to their respective objectives.
Second, we factorize the variation process to distinguish reusable harness
mechanisms from environment-specific adaptation. General-purpose and
environment-focused modifications are evolved through different operators and
dynamically composed at inference time. Figure~\ref{fig:method-overview}
illustrates the complete evolution loop.
\paragraph{Search Space: The Harness as the Genome}
\label{sec:what}

Unlike program-evolution methods that optimize a compact code artifact, the
individual in \methodname is an \emph{entire agent harness}. We freeze only a
minimal interface required by the evaluator; the remaining
harness---including prompts, tools, skills, control logic, verification
mechanisms, and orchestration---constitutes the evolvable genome.

Concretely, the search space contains three major classes of components,
$\mathcal{K}=\{\text{tools}, \text{skills}, \text{logic}\}$, where logic
also covers orchestration.
\emph{Tools} provide executable capabilities available inside the agent loop,
including environment interaction and verification utilities. 
\emph{Skills} encode reusable domain knowledge, such as debugging procedures or
repair strategies, that can be retrieved when relevant. Finally,
\emph{logic and orchestration} determine how these capabilities are used,
including the system prompt, agent loop, verification gates, and workflow
connecting individual reasoning and execution stages. Because the proposer
operates directly over the harness repository, evolution may modify not only
local instructions but also the high-level structure of the agent. The initial harness $H_{0}$ is a lightweight but functional ReAct-style base (details in Appendix~\ref{app:init}).

\subsection{Multi-Objective Evolution by Decomposition}
\label{sec:moead}

Directly maintaining a single ``best'' harness is ill-defined under multiple
objectives for two reasons: (1) evolutionary search must maintain sufficient
population diversity to avoid premature convergence; and (2) different
harnesses may realize distinct Pareto-optimal trade-offs. We therefore
instantiate harness evolution using a decomposition strategy based on
MOEA/D~\citep{li2009moeadde}.

\paragraph{Decomposition.}
Let
$
    \Lambda
    =
    \left\{
        \boldsymbol{\lambda}^{1},
        \ldots,
        \boldsymbol{\lambda}^{N}
    \right\},
    \lambda^{i}_{j} \geq 0,
    \sum_{j=1}^{m} \lambda^{i}_{j} = 1,
$
denote a set of $N$ weight vectors spanning the $m$-objective simplex.
In our bi-objective setting ($m=2$), the weight vectors are chosen to be
uniformly distributed along the security--functionality trade-off. Each
$\boldsymbol{\lambda}^{i}$ defines a scalar optimization subproblem and is
associated with an incumbent harness $H_i$. Collectively, the $N$ subproblems
cover different objective preferences and maintain a finite set of
representative solutions that approximate the Pareto frontier.

We maintain an ideal point
$\mathbf{z}^{*}=(z_1^{*},\ldots,z_m^{*})$, where
$z_j^{*}=\max_{H}\hat f_j(H)$ is the best value of objective $j$ observed over
all harnesses evaluated so far. Given a weight vector
$\boldsymbol{\lambda}^{i}\in\Lambda$, we define the corresponding subproblem
using the weighted Tchebycheff scalarization
\begin{equation}
    g^{\mathrm{te}}
    \bigl(
        H \mid \boldsymbol{\lambda}^{i},\mathbf{z}^{*}
    \bigr)
    =
    \max_{j\in\{1,\ldots,m\}}
    \lambda^{i}_{j}
    \left|
        z^{*}_{j}-\hat f_j(H)
    \right|,
    \label{eq:tchebycheff}
\end{equation}
where lower values indicate a smaller weighted deviation from the ideal point.
Conceptually, the solution associated with subproblem $i$ is
\begin{equation}
    H_i^{*}
    =
    \arg\min_{H\in\mathcal{H}}
    g^{\mathrm{te}}
    \bigl(
        H \mid \boldsymbol{\lambda}^{i},\mathbf{z}^{*}
    \bigr),
    \label{eq:subproblem}
\end{equation}
and during evolution we maintain an incumbent $H_i$ as the current
approximation to $H_i^{*}$.

This decomposition is fundamentally different from optimizing a single
weighted-sum objective. 
Importantly, \textbf{the scalarization in
Eq.~\ref{eq:tchebycheff} is \emph{local to a subproblem}:} \methodname does not
replace the multi-objective problem with one globally authoritative scalar
reward. Different subproblems preserve different objective preferences, so a
harness favored by a security-oriented subproblem need not compete directly
with one favored by a functionality-oriented subproblem.

\textbf{Neighborhood evolution and replacement.}
For each subproblem $i$, we define a neighborhood $B(i)$ as the set of the
$T$ weight vectors in $\Lambda$ closest to $\boldsymbol{\lambda}^{i}$ under
Euclidean distance. Variation is single-parent: the child for subproblem $i$
is produced from its own incumbent $H_i$. The neighborhood is used in two
places. First, if a subproblem stalls, meaning its parent and failure digest are
unchanged from the previous generation, we breed instead from the neighbor with the best
Tchebycheff value under $\boldsymbol{\lambda}^{i}$. Second, the neighborhood
defines which incumbents the child may replace (below). Both rely on the
intuition that nearby objective preferences benefit from similar harness
designs. The proposer never receives more than one full parent repository, which
keeps its context bounded.

Suppose variation produces a child harness $H_c$ for subproblem $i$. After
evaluating $H_c$ on all objectives and updating the ideal point
$\mathbf{z}^{*}$, we consider $H_c$ as a replacement candidate for each
neighboring subproblem $l\in B(i)$:
\begin{equation}
    H_l \leftarrow H_c
    \quad \text{if} \quad
    g^{\mathrm{te}}
    \bigl(
        H_c \mid \boldsymbol{\lambda}^{l}, \mathbf{z}^{*}
    \bigr)
    <
    g^{\mathrm{te}}
    \bigl(
        H_l \mid \boldsymbol{\lambda}^{l}, \mathbf{z}^{*}
    \bigr).
    \label{eq:replacement}
\end{equation}
Thus, a candidate discovered for one subproblem may improve several nearby
objective preferences, while incumbents associated with substantially different
trade-offs are preserved. Standard MOEA/D bounds the number of incumbents one
child may replace by $n_r\le T$~\citep{li2009moeadde}. We cap replacement at
$n_r=2$, so a single child can replace at most two neighbouring incumbents.

\textbf{Objective-induced diversity and bounded proposer context.}
This decomposition is particularly well suited to repository-scale harness
evolution. Existing evolutionary methods~\citep{novikov2025alphaevolve,wang2025thetaevolve}
often maintain diversity using explicit artifact-level distance measures,
which are more natural when each candidate is a compact program for a single
task. For repository-scale harnesses, however, structural distance is a poor
proxy for behavioral diversity: two repositories may implement nearly identical
behavior with very different code organization, while small changes to a prompt,
verification gate, or orchestration rule can substantially alter behavior.
Designing a reliable hand-crafted diversity metric is therefore difficult when
candidates vary jointly over prompts, tools, control flow, verification, and
workflow structure.
In \methodname, each incumbent is preserved because it is the best found so far
for a distinct objective preference, not because it is structurally distant
under a hand-designed metric. This makes diversity of \emph{selection pressure}
hold by construction. Diversity of \emph{behavior} does not follow automatically.
It also requires the objectives to trade off and the LLM operators to make
edits that change behavior rather than wording. Measuring this is itself an open problem for
self-coding harnesses. Our offline analysis (App.~\ref{sec:diversity}) finds that
standard program-diversity metrics (token, line and AST distance, and edit labels)
barely track behaviour: code distance and objective distance are essentially
uncorrelated. The metrics also cannot see diversity that is decided at run time,
such as the distinct per-task DAGs planned by the orchestrator. Among them, AST
distance is the most informative, since it ignores cosmetic rewrites. We
therefore treat diversity as something to monitor, not something decomposition
guarantees. \textbf{The population size is also bounded by the number of
subproblems}, avoiding an ever-growing archive and reducing the evolutionary
context that must be maintained.

\subsection{Factorized Harness Evolution}
\label{sec:operators}


For a target subproblem $i$, the proposer
receives a single parent harness, the incumbent $H_i$ (or a neighbor when the
subproblem stalls), the objective
preference $\boldsymbol{\lambda}^{i}$, and the execution feedback $\mathcal{E}_i$ collected
from running that parent. The feedback contains aggregate objective
statistics together with sampled execution artifacts, such as compiler errors,
failing tests, and crash traces. Before modifying the harness, the proposer
first diagnoses the observed failures and formulates a concrete hypothesis for
improvement under the current objective preference.

\textbf{Why factorize harness evolution?}
A repository-scale harness contains heterogeneous components, including tools,
skills, prompts, verification logic, control flow, and orchestration, that play
different roles in agent behavior. An observed failure often suggests a change
to only a subset of these components. Treating the entire repository as a
single rewrite target can therefore couple an intended repair with unrelated
changes elsewhere in the harness. More importantly, because evolutionary
feedback is collected from a particular benchmark, unrestricted rewriting does
not explicitly distinguish reusable mechanisms from adaptations that exploit
properties specific to the current benchmark.

\textbf{Factorized harness representation.} We therefore factorize harness evolution along two complementary dimensions:
\emph{component type} and \emph{adaptation scope}. Component factorization
localizes an update to a semantically meaningful part of the harness, while
scope factorization provides an explicit representation for reusable and
environment-specific components. This organization serves as an inductive bias;
we do not assume that either decomposition is unique or theoretically exact.
Given execution feedback $\mathcal{E}_i$, the proposer first diagnoses the observed
failure and formulates an improvement hypothesis:
$
d_i = \mathcal{Q}(\mathcal{E}_i),
h_i = \mathcal{G}(d_i, \boldsymbol{\lambda}^{i}),
$
where $\mathcal{Q}$ and $\mathcal{G}$ denote the diagnosis and hypothesis
steps of the proposer, $d_i$ denotes the failure diagnosis, and $h_i$ denotes
the proposed improvement hypothesis.
Based on this hypothesis, the proposer selects the component class to modify,
$
k_i
=
\pi_{\mathrm{comp}}(h_i)
\in
\mathcal{K},
$
where $\mathcal{K}$ is the set of component classes defined in
\S\ref{sec:what}.
The proposer then assigns the scope of the modification through a
component-specific scope assignment $\mathcal{R}_{k_i}$:
\begin{equation} 
\label{eq:distinction}
s_i
=
\mathcal{R}_{k_i}
\left(
H_i,
h_i
\right)
\in
\Sigma,
\end{equation}
where
$
\Sigma
=
\{
\mathrm{gen},
\mathrm{env}
\}.
$
Here, $k_i$ determines \emph{what component to modify}, while $s_i$
determines \emph{where the resulting modification should reside}. gen refers to a general harness component, and env means environment-specific. 
The corresponding component-scope update is then applied to produce the child
harness of \S\ref{sec:moead}:
$
H_c
=
\mathcal{O}_{k_i,s_i}
\left(
H_i;
h_i
\right).
$
Here, $\mathcal{O}_{k_i,s_i}$ is a f\textbf{actorized variation operator} that directly modifies the harness. The complete operator details are provided in
Appendix~\ref{app:factorized-evolution}.
The distinction in Eq.~\ref{eq:distinction} is operational rather than
semantic. In the implementation, components are stored under explicit general
or benchmark-specific scopes. For example, tools and skills may each contain a
general implementation together with benchmark-specific variants. The proposer
therefore makes two decisions when modifying the harness: which component to
change, and whether the resulting change should reside in the general scope or
in an environment-specific scope. This assignment is not fixed and may be
revised during evolution.

\textbf{Generalization as cross-scope reorganization.}
The boundary between general and environment-specific components is itself
allowed to evolve. We introduce a dedicated \texttt{generalize} operator that
inspects the current harness organization and reorganizes components across the
two scopes when their existing placement introduces unnecessary specialization. Implementation details
for this operator are provided in Appendix~\ref{app:generalize}.

\textbf{Validation and recovery.}
Every proposed child is validated before objective evaluation in three stages. The details are described in
Appendix~\ref{app:validation-recovery}.

\subsection{Dynamic Composition at Inference}
\label{sec:composition}

Factorized evolution produces a pool of reusable components
$\mathcal{H}^{\mathrm{gen}}$ together with environment-specific components
$\mathcal{H}^{\mathrm{env}}_{e}$. We do not execute the complete union of these
components for every task. Even within the same benchmark, different tasks may
benefit from different tools, skills, verification procedures, or control
strategies.
At inference time, a dedicated composition agent constructs the active harness
conditioned on both the current environment and the current task. Given an
environment $e$, a task $x$, and the available component pools, the composition
agent generates a task-specific DAG,
\begin{equation}
    \mathcal{A}_{e,x}
    =
    \pi_{\phi}
    \left(
        e,
        x,
        \mathcal{H}^{\mathrm{gen}},
        \mathcal{H}^{\mathrm{env}}_{e}
    \right),
    \label{eq:component-selection}
\end{equation}
where $\pi_{\phi}$ denotes the composition policy and
$\mathcal{A}_{e,x}$ contains the components activated for the current task.
The executable harness is then instantiated as
$
    H^{(e,x)}
    =
    \mathcal{C}\left(\mathcal{A}_{e,x}\right),
$
where $\mathcal{C}$ denotes the runtime composition procedure.
The composition agent does not modify the
evolved components; it only determines which components to activate for the
current environment and task. This allows general mechanisms to remain
available across environments without being invoked unconditionally, while
environment-specific mechanisms are loaded only when they are relevant to the
current task. Details are provided in
Appendix~\ref{app:dynamic-composition}.

\section{Experiments}
\label{sec:experiments}

For each benchmark, we partition tasks into \emph{evolve},
\emph{held-out-ID}, and \emph{held-out-OOD} splits. The evolve split provides
the execution feedback used during harness evolution, whereas both held-out
splits remain inaccessible to the proposer and selection procedure.
For SecRepoBench, we use 30/30/30 tasks; for BaxBench, we use 28/28/28 tasks
and evolve a separate harness population on each benchmark. We report
\textbf{secure-pass@1}, defined as the conjunction of functional and security
success, $F \wedge S$.  Unless otherwise specified, both the harness proposer and the
underlying task-solving agent use \texttt{gemini-3.6-flash}. Appendix~\ref{sec:exp-setup} describes the benchmarks in detail.

\subsection{Main comparison against self-evolution baselines}
\label{sec:exp-main}
\begin{figure}[t]
\centering
\begin{subfigure}{0.8\linewidth}
  \centering
  \includegraphics[width=\linewidth]{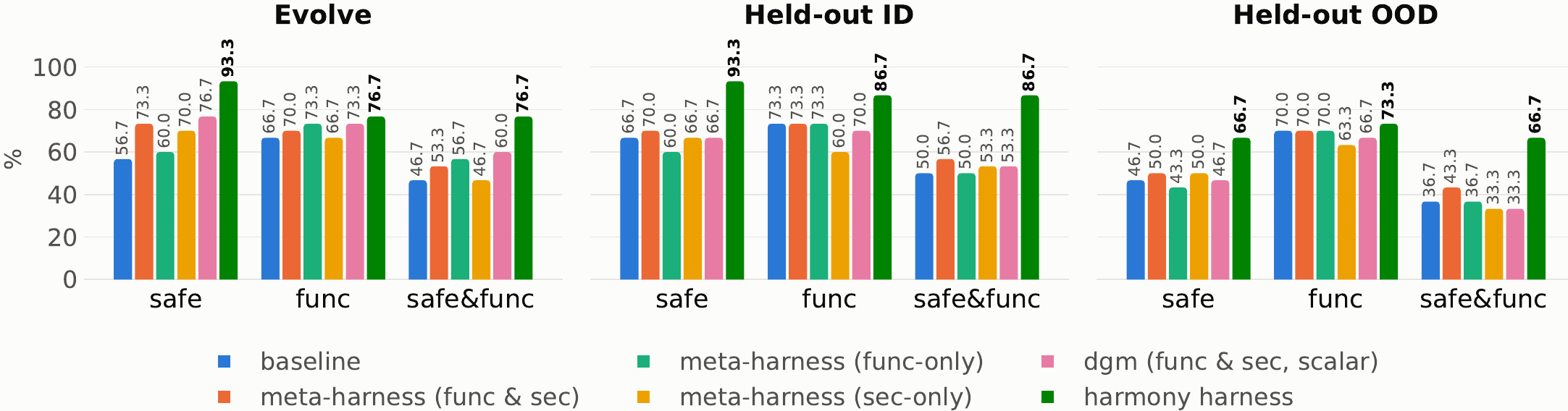}
\end{subfigure}

\vspace{2pt}
\begin{subfigure}{0.8\linewidth}
  \centering
  \includegraphics[width=\linewidth]{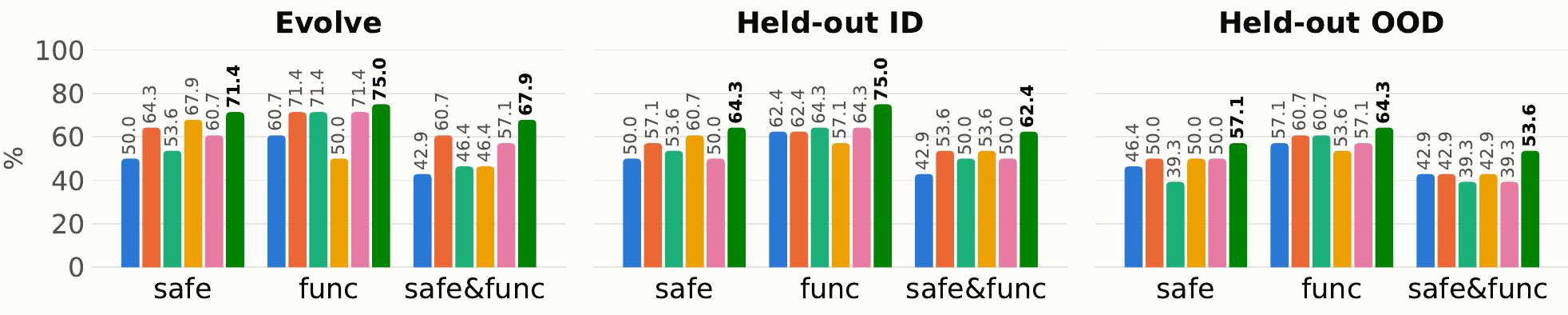}
\end{subfigure}
\caption{Evolution results per split on (a) SecRepoBench and (b) BaxBench. }
\label{fig:secrepo-bars}
\vspace{-0.5cm}
\end{figure}

We first compare six evolution regimes under an identical seed harness
and grader. \textbf{Baseline} is the un-evolved harness. The three
\textbf{meta-harness}~\citep{lee2026meta} variants are its monolithic
``summarize$\to$rewrite'' loop driven by (i) functionality \emph{and} security,
(ii) functionality only, or (iii) security only. \textbf{DGM}~\citep{zhang2026darwin}
scalarizes the two objectives into a single reward. \textbf{HH} (\methodname) is ours.
Each method is evolved on the evolve split only and then evaluated, without
further evolution, on both held-out splits. Figure~\ref{fig:secrepo-bars} summarizes the outcome per split on the two benchmarks;
the underlying numbers are tabulated in Tables~\ref{tab:evolve-secrepo}
and~\ref{tab:bax-main} in Appendix~\ref{sec:appendix-tables}.

Table~\ref{tab:retention} crystallizes the central failure mode we study: baseline methods improve security on the optimization distribution yet relinquish most or all of those gains under distribution shift—with larger in-distribution gains often corresponding to larger reversals, as exemplified by DGM, which achieves the strongest SecRepoBench security improvement among the baselines during evolution but retains none of it out-of-distribution—motivating our treatment of safety as an explicit optimization axis evaluated beyond the evolve distribution rather than as a scalar reward measured only in-distribution. HH is
best on every split of both benchmarks: $76.7$ F\&S on the SecRepoBench evolve
split, $86.7$ on held-out IN (vs.\ $50.0$ for baseline), and $66.7$ on held-out
OUT (vs.\ $36.7$), and an average F\&S of $61.3$ on BaxBench
against $42.9$ for baseline and $52.4$ for the strongest meta-harness variant.
Notably the largest \emph{relative} margins are out-of-distribution, which is
what factorizing the harness into general and environment-specific components is
meant to achieve.

\begin{table}[t]
\centering
\small
\setlength{\tabcolsep}{5pt}
\caption{\textbf{How much evolved security survives a distribution shift.}
$\Delta_{\text{evolve}}$ is a method's security gain over the un-evolved
baseline on the split it evolved on; $\Delta_{\text{OUT}}$ is its gain over the
same baseline on held-out-OUT; \emph{retained} $=
\Delta_{\text{OUT}}/\Delta_{\text{evolve}}$. Mean over 3 runs.}
\label{tab:retention}
\begin{tabular}{@{}l ccc ccc@{}}
\toprule
& \multicolumn{3}{c}{SecRepoBench} & \multicolumn{3}{c}{BaxBench} \\
\cmidrule(lr){2-4}\cmidrule(lr){5-7}
Method & $\Delta_{\text{evolve}}$ & $\Delta_{\text{OUT}}$ & retained
       & $\Delta_{\text{evolve}}$ & $\Delta_{\text{OUT}}$ & retained \\
\midrule
Meta-harness (func-only)    & $+3.3$  & $-3.4$  & $-103\%$ & $+3.6$  & $-7.1$  & $-197\%$ \\
Meta-harness (sec-only)     & $+13.3$ & $+3.3$  & $25\%$   & $+17.9$ & $+3.6$  & $20\%$ \\
Meta-harness (func \& sec)  & $+16.6$ & $+3.3$  & $20\%$   & $+14.3$ & $+3.6$  & $25\%$ \\
DGM (func \& sec, scalar)   & $+20.0$ & $+0.0$  & ${0\%}$ & $+10.7$ & $+3.6$ & $34\%$ \\
\midrule
\methodname (func \& sec)                  & $+36.6$ & $+20.0$ & $\mathbf{55\%}$
                            & $+21.4$ & $+10.7$ & $\mathbf{50\%}$ \\
\bottomrule
\end{tabular}
\vspace{-0.1cm}
\end{table}

\begin{table}[t]
\centering
\small
\caption{Evolution strategies on SecRepoBench.
Func/Sec/F\&S in \%. Mean over 3 runs.}
\label{tab:strategy}
\begin{tabular}{@{}llccc@{}}
\toprule
Strategy & Acceptance rule & Func & Sec & F\&S  \\
\midrule
Static (no evolution)            & ---                                    & 66.7 & 56.7& 46.7 \\
Functionality-only               & accept if $F\!\uparrow$                & 73.3 & 60.0 & 56.7\\
Safety-only                      & accept if $\text{viol.}\!\downarrow$   & 70.0 & 66.7 & 46.7\\
\midrule
Sequential (func$\to$sec, sec$\to$func) & one objective per generation, then the other & 70.0 & 76.7 & 70.0  \\
Prompt-merge variants            & merge func \& safety prompts           & 66.7 & 60.0 & 50.0  \\
Joint-weighted ($\alpha\!\in\!\{.25,.5,.75\}$) & $\alpha F\!+\!(1\!-\!\alpha)S$ & 70.0 & 73.3 & 70.0   \\
Constrained (hard $\epsilon$) & \cite{achiam2017cpo}         & 70.0 & 60.0 & 60.0  \\
Separated func/safety combination     & dual harness combination                     & 60.0 & 60.0 & 53.3  \\
\methodname        & Tchebycheff (Eq.~\ref{eq:replacement})           & \textbf{76.7} & \textbf{93.3} & \textbf{76.7} \\
\bottomrule
\end{tabular}
\vspace{-0.5cm}
\end{table}

\textbf{Comparison with possible multi-objective evolving strategies.}
\label{sec:exp-strategy} 
In Table~\ref{tab:strategy}, we compare with various baselines with the same initial harness, evolve split, token budget, and settings, through the alternative ways of combining the two objectives
(sequential, weighted sum, hard constraint), and a variant that evolves the two
harnesses separately and combines them afterwards.
Four readings stand out. \textbf{(1) Single-objective evolution does not solve
the joint problem}: functionality-only lifts F\&S from the static $46.7$ to
$56.7$ but leaves security at $60.0$, while safety-only buys $10.0$ points of
security ($56.7\!\to\!66.7$) that never convert into a single additional
secure-pass---its F\&S stays at the static $46.7$.
\textbf{(2) With factored evolving and dynamic orchestration held fixed, the acceptance rule matters.} Under these matched conditions, Tchebycheff replacement reaches 76.7 F\&S, against 70.0 for both the weighted sum and sequential evolution, 60.0 for the hard constraint, and 53.3 for separately evolved harnesses. Factored evolving and dynamic orchestration are also necessary: the component ablation in Table~\ref{tab:main-ablation} in App.~\ref{sec:appendix-tables} adds them one at a time on top of Tchebycheff joint evolution, lifting F\&S from $63.3$ to $66.7$ and then to $76.7$, so each part of the method contributes and only the full design reaches the headline result.
\textbf{(3) A hard feasibility constraint acts as a target rather than a
direction.} Constrained acceptance holds functionality
at $70.0$ but leaves security at $60.0$, only $3.3$ above never evolving at all.
This is the behaviour an $\epsilon$-threshold rule predicts: it rejects anything
below the bar and is indifferent among everything above it, so evolution stops
improving safety the moment safety becomes feasible, whereas Tchebycheff
replacement keeps pulling on whichever objective has the largest weighted shortfall.
\textbf{(4) Generalization: real execution feedback beats pattern-matching.}
\label{sec:exp-generalization}
The central positive result is on \emph{held-out} tasks
(Table~\ref{tab:heldout} in App.~\ref{sec:appendix-tables}). Per-generation bests of the execution-feedback variant
improve combined F\&S from the baseline's $43.3$ to $58.3$, whereas the
baseline \emph{regresses} to $35.0$. The gain concentrates on
\textbf{held-out OUT} (unseen crash types), where real exploit feedback
lets a harness stop \emph{unfamiliar} vulnerabilities instead
of pattern-matching known ones; the proposed method reaches $66.7$ out-of-distribution
and $76.7$ combined. Consulting ground truth, not memorizing a distribution, is
what generalizes.


\textbf{Dynamic orchestration vs.\ a fixed pipeline.}
\label{sec:exp-orchestration}
Table~\ref{tab:orch} in App.~\ref{sec:appendix-tables}
reports dynamic-workflow vs.\ linear-pipeline on BaxBench at matched harness and
budget.
Per-task planning is worth $+10.7$ points of secure-pass and $+3.6$ of
functional-pass for an $8\%$ token overhead ($58$k to $63$k) and $1.2$ extra
steps on average---the orchestrator spends more only where it helps.
Figure~\ref{fig:bax-dag} in the appendix shows what it actually plans: the seven distinct DAG
shapes emitted over $28$ BaxBench tasks. The modal plan ($39\%$ of tasks) is a
five-node fan-out that runs \texttt{build\_verify}, \texttt{endpoint\_check}, and
\texttt{security\_audit} in parallel and joins into a single \texttt{edit\_fix};
harder tasks get $6$--$8$ nodes with conditional \texttt{when}-guarded
\texttt{shell\_probe} and repair loops that only fire if an earlier check fails.
A fixed pipeline must either pay for those extra stages on every task or never
have them; the orchestrator recovers both ends of that trade-off, which is where
the F\&S gap comes from.

\subsection{Cross-benchmark transfer}
\label{sec:exp-transfer}

\begin{figure}[t]
\centering
\includegraphics[width=0.8\linewidth]{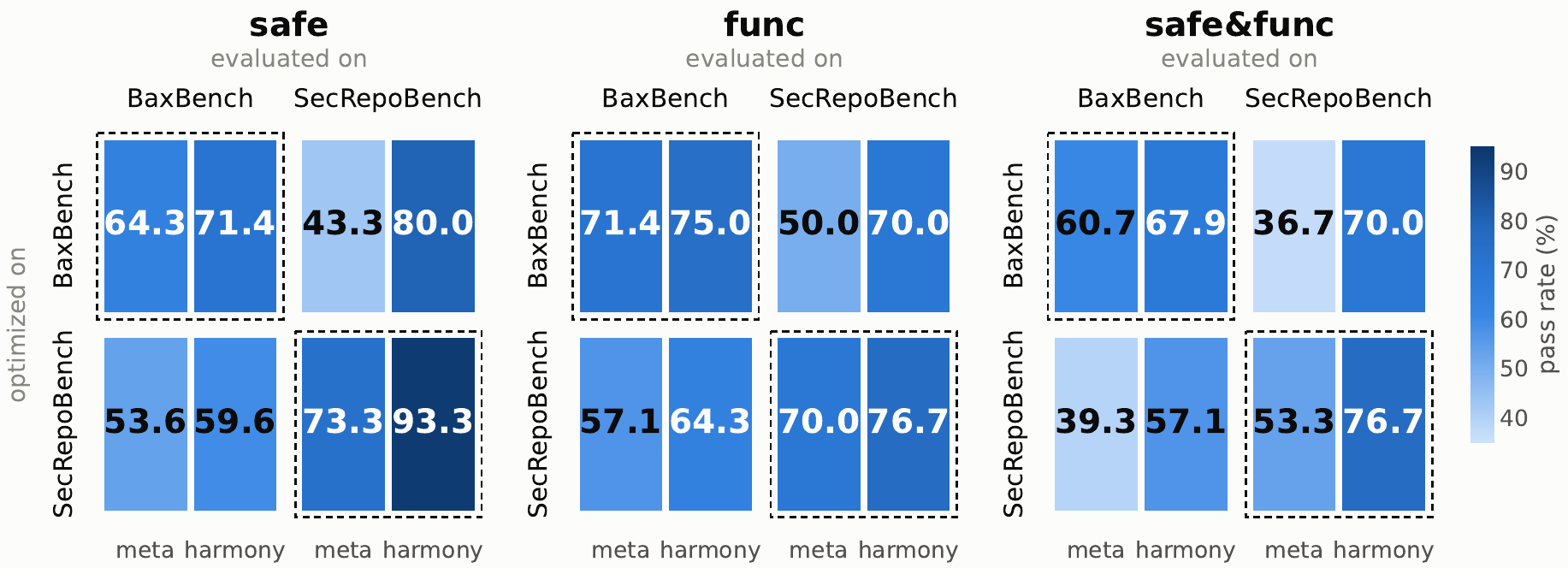}
\caption{Cross-benchmark transfer. Each cell is one
(optimized-on $\times$ evaluated-on) pair, split into meta-harness (left) and
\methodname (right); dashed outlines mark in-domain cells, solid outlines zero-shot
transfer. }
\label{fig:transfer-matrix}
\vspace{-0.7cm}
\end{figure}

We evaluate each evolved harness zero-shot on the other benchmark's search
split (Figure~\ref{fig:transfer-matrix}, Table~\ref{tab:transfer} in App.~\ref{sec:appendix-tables}); the target
benchmark is never seen during evolution. Meta-harness transfers poorly in both
directions: it drops to $39.3$ F\&S on BaxBench (baseline $42.9$) and to $36.7$
on SecRepoBench (baseline $46.7$). \methodname improves over the baseline in
both directions, reaching $57.1$ on BaxBench ($+14.2$) and $70.0$ on
SecRepoBench ($+23.3$), close to the $76.7$ it obtains when evolved in-domain.
We attribute this to the separation between general and benchmark-specific
components: only the general scope is loaded on the target benchmark, so
benchmark-specific fixes do not carry over.

\textbf{CWEval.} In Table~\ref{tab:cweval-transfer} in App.~\ref{sec:appendix-tables}, the harness evolved on BaxBench is run, again with no
further evolution, on CWEval~\citep{peng2025cweval}, which changes not just the
benchmark but the \emph{task format}: generate a single function from a
natural-language spec, graded by functional tests and a separate security oracle,
over $357$ instances in five languages. \textbf{(1) It greatly improves security, which suggests the harness transfers to a benchmark it has never seen.} \textbf{(2) It holds in every
language, including ones BaxBench does not contain.} Per-language F\&S
improves everywhere, from $+10.1$ (JavaScript) to $+19.3$ (Go). C and C++ are not
among BaxBench's six languages, yet they are $156$ of the $357$ instances and
gain $+13.9$ and $+14.3$, which suggests the generalization of the \methodname.
\textbf{(3) Functionality pays for part of it.} Functional pass rate falls on
three of the five languages (C++, JavaScript, Python), and it gives up $2.0$ points of functionality for $15.5$ points of security and $14.3$ points of F\&S.



\section{Conclusion}
Self-evolving agents optimized for a single objective quietly sacrifice the
other, and specialize to one distribution at the expense of safety on others. We
recast self-evolution as multi-objective search over an entire agent harness,
with safety kept as a \emph{separate axis} of evolution rather than a reward term
to be averaged away. In secure code generation---where
both objectives are grounded in real execution---joint, harness-level evolution
improves security and functionality together and, with execution-grounded
feedback, generalizes to unseen vulnerabilities, while naive feedback-driven
evolution falls back. The factorized general and environment-specific harness further improves the in-domain and cross-benchmark generalization. 

\subsection*{AI use statement}
In this work, we used generative AI tools for editing (grammar, spelling, and word
choice), for drafting sections of the paper, for implementing standard methods in
our codebase, and for retrieval and discovery of related work: candidate references were
surfaced with AI assistance and every one of them was then read and verified against its
primary source by an author before being cited or described. We did not use generative AI
tools to generate research ideas, or to generate or fabricate experimental results. All
AI-assisted text was reviewed and edited by the authors; all AI-assisted code was read, executed, and tested by at least two
authors, and every number reported in this paper comes from a run we performed
ourselves. Generative AI was not used as a subject of study, as a research participant,
or as an evaluator of our results. We take responsibility for the final content of this
work, including all text, claims, and artifacts produced with the aid of generative AI.

\bibliographystyle{iclr2027_conference}
\bibliography{references}

@article{novikov2025alphaevolve,
  title         = {{AlphaEvolve}: A Coding Agent for Scientific and Algorithmic Discovery},
  author        = {Novikov, Alexander and V{\~u}, Ng{\^a}n and Eisenberger, Marvin and
                   Dupont, Emilien and Huang, Po-Sen and Wagner, Adam Zsolt and
                   Shirobokov, Sergey and Kozlovskii, Borislav and Ruiz, Francisco J. R. and
                   Mehrabian, Abbas and Kumar, M. Pawan and See, Abigail and
                   Chaudhuri, Swarat and Holland, George and Davies, Alex and
                   Nowozin, Sebastian and Kohli, Pushmeet and Balog, Matej},
  journal       = {arXiv preprint arXiv:2506.13131},
  year          = {2025},
  eprint        = {2506.13131},
  archivePrefix = {arXiv},
  primaryClass  = {cs.AI},
  url           = {https://arxiv.org/abs/2506.13131}
}

@article{wang2025thetaevolve,
  title         = {{ThetaEvolve}: Test-time Learning on Open Problems},
  author        = {Wang, Yiping and Su, Shao-Rong and Zeng, Zhiyuan and Xu, Eva and
                   Ren, Liliang and Yang, Xinyu and Huang, Zeyi and He, Xuehai and
                   Ma, Luyao and Peng, Baolin and Cheng, Hao and He, Pengcheng and
                   Chen, Weizhu and Wang, Shuohang and Du, Simon Shaolei and Shen, Yelong},
  journal       = {arXiv preprint arXiv:2511.23473},
  year          = {2025},
  eprint        = {2511.23473},
  archivePrefix = {arXiv},
  primaryClass  = {cs.LG},
  url           = {https://arxiv.org/abs/2511.23473}
}

@article{li2009moeadde,
  title   = {Multiobjective Optimization Problems With Complicated Pareto Sets, {MOEA/D} and {NSGA-II}},
  author  = {Li, Hui and Zhang, Qingfu},
  journal = {IEEE Transactions on Evolutionary Computation},
  volume  = {13},
  number  = {2},
  pages   = {284--302},
  year    = {2009},
  doi     = {10.1109/TEVC.2008.925798}
}

@article{wang2023voyager,
  title     = {Voyager: An Open-Ended Embodied Agent with Large Language Models},
  author    = {Wang, Guanzhi and Xie, Yuqi and Jiang, Yunfan and Mandlekar, Ajay and Xiao, Chaowei and Zhu, Yuke and Fan, Linxi and Anandkumar, Anima},
  journal   = {Transactions on Machine Learning Research},
  year      = {2024},
  eprint    = {2305.16291},
  archivePrefix = {arXiv},
  url       = {https://arxiv.org/abs/2305.16291}
}

@inproceedings{shinn2023reflexion,
  title     = {Reflexion: Language Agents with Verbal Reinforcement Learning},
  author    = {Shinn, Noah and Cassano, Federico and Gopinath, Ashwin and Narasimhan, Karthik and Yao, Shunyu},
  booktitle = {Advances in Neural Information Processing Systems (NeurIPS)},
  year      = {2023},
  eprint    = {2303.11366},
  archivePrefix = {arXiv},
  url       = {https://arxiv.org/abs/2303.11366}
}

@inproceedings{sun2023adaplanner,
  title     = {AdaPlanner: Adaptive Planning from Feedback with Language Models},
  author    = {Sun, Haotian and Zhuang, Yuchen and Kong, Lingkai and Dai, Bo and Zhang, Chao},
  booktitle = {Advances in Neural Information Processing Systems (NeurIPS)},
  year      = {2023},
  eprint    = {2305.16653},
  archivePrefix = {arXiv},
  url       = {https://arxiv.org/abs/2305.16653}
}

@inproceedings{hu2024evomac,
  title     = {Self-Evolving Multi-Agent Collaboration Networks for Software Development},
  author    = {Hu, Yue and Cai, Yuzhu and Du, Yaxin and Zhu, Xinyu and Liu, Xiangrui and Yu, Zijie and Hou, Yuchen and Tang, Shuo and Chen, Siheng},
  booktitle = {International Conference on Learning Representations (ICLR)},
  year      = {2025},
  eprint    = {2410.16946},
  archivePrefix = {arXiv},
  url       = {https://arxiv.org/abs/2410.16946}
}

@inproceedings{zhang2025aflow,
  title={Aflow: Automating agentic workflow generation},
  author={Zhang, Jiayi and Xiang, Jinyu and Yu, Zhaoyang and Teng, Fengwei and Chen, Xionghui and Chen, Jiaqi and Zhuge, Mingchen and Cheng, Xin and Hong, Sirui and Wang, Jinlin and others},
  booktitle={International Conference on Learning Representations},
  volume={2025},
  pages={34040--34077},
  year={2025}
}

@inproceedings{sinha2024survival,
  title={Survival of the Safest: Towards Secure Prompt Optimization through Interleaved Multi-Objective Evolution},
  author={Sinha, Ankita and Cui, Wendi and Das, Kamalika and Zhang, Jiaxin},
  booktitle={Proceedings of the 2024 Conference on Empirical Methods in Natural Language Processing: Industry Track},
  pages={1016--1027},
  year={2024}
}

@inproceedings{agrawal2026gepa,
  title     = {{GEPA}: Reflective Prompt Evolution Can Outperform Reinforcement Learning},
  author    = {Agrawal, Lakshya A and Tan, Shangyin and Soylu, Dilara and Ziems, Noah and Khare, Rishi and Opsahl-Ong, Krista and Singhvi, Arnav and Shandilya, Herumb and Ryan, Michael J. and Jiang, Meng and Potts, Christopher and Sen, Koushik and Dimakis, Alexandros G. and Stoica, Ion and Klein, Dan and Zaharia, Matei and Khattab, Omar},
  booktitle = {International Conference on Learning Representations (ICLR)},
  year      = {2026},
  eprint    = {2507.19457},
  archivePrefix = {arXiv},
  url       = {https://arxiv.org/abs/2507.19457}
}

@inproceedings{hu2024adas,
  title     = {Automated Design of Agentic Systems},
  author    = {Hu, Shengran and Lu, Cong and Clune, Jeff},
  booktitle = {International Conference on Learning Representations (ICLR)},
  year      = {2025},
  eprint    = {2408.08435},
  archivePrefix = {arXiv},
  url       = {https://arxiv.org/abs/2408.08435}
}

@inproceedings{tang2024agentpro,
  title     = {Agent-Pro: Learning to Evolve via Policy-Level Reflection and Optimization},
  author    = {Zhang, Wenqi and Tang, Ke and Wu, Hai and Wang, Mengna and Shen, Yongliang and Hou, Guiyang and Tan, Zeqi and Li, Peng and Zhuang, Yueting and Lu, Weiming},
  booktitle = {Proceedings of the 62nd Annual Meeting of the Association for Computational Linguistics (Volume 1: Long Papers)},
  pages     = {5348--5375},
  year      = {2024},
  eprint    = {2402.17574},
  archivePrefix = {arXiv},
  url       = {https://aclanthology.org/2024.acl-long.292/}
}

@inproceedings{levy2024stwebagent,
  title     = {ST-WebAgentBench: A Benchmark for Evaluating Safety and Trustworthiness in Web Agents},
  author    = {Levy, Ido and Wiesel, Ben and Marreed, Sami and Oved, Alon and Yaeli, Avi and Mashkif, Nir and Shlomov, Segev},
  booktitle = {International Conference on Learning Representations (ICLR)},
  year      = {2026},
  eprint    = {2410.06703},
  archivePrefix = {arXiv},
  url       = {https://arxiv.org/abs/2410.06703}
}

@inproceedings{debenedetti2024agentdojo,
  title     = {AgentDojo: A Dynamic Environment to Evaluate Prompt Injection Attacks and Defenses for {LLM} Agents},
  author    = {Debenedetti, Edoardo and Zhang, Jie and Balunovi{\'c}, Mislav and Beurer-Kellner, Luca and Fischer, Marc and Tram{\`e}r, Florian},
  booktitle = {Advances in Neural Information Processing Systems (NeurIPS), Datasets and Benchmarks Track},
  year      = {2024},
  eprint    = {2406.13352},
  archivePrefix = {arXiv},
  url       = {https://arxiv.org/abs/2406.13352}
}

@article{yin2024safeagentbench,
  title     = {SafeAgentBench: A Benchmark for Safe Task Planning of Embodied {LLM} Agents},
  author    = {Yin, Sheng and Pang, Xianghe and Ding, Yuanzhuo and Chen, Menglan and Bi, Yutong and Xiong, Yichen and Huang, Wenhao and Xiang, Zhen and Shao, Jing and Chen, Siheng},
  journal   = {arXiv preprint arXiv:2412.13178},
  year      = {2024},
  url       = {https://arxiv.org/abs/2412.13178}
}

@inproceedings{ruan2024toolemu,
  title     = {Identifying the Risks of {LM} Agents with an {LM}-Emulated Sandbox},
  author    = {Ruan, Yangjun and Dong, Honghua and Wang, Andrew and Pitis, Silviu and Zhou, Yongchao and Ba, Jimmy and Dubois, Yann and Maddison, Chris J. and Hashimoto, Tatsunori},
  booktitle = {International Conference on Learning Representations (ICLR)},
  year      = {2024},
  eprint    = {2309.15817},
  archivePrefix = {arXiv},
  url       = {https://arxiv.org/abs/2309.15817}
}

@inproceedings{andriushchenko2024agentharm,
  title     = {AgentHarm: A Benchmark for Measuring Harmfulness of {LLM} Agents},
  author    = {Andriushchenko, Maksym and Souly, Alexandra and Dziemian, Mateusz and Duenas, Derek and Lin, Maxwell and Wang, Justin and Hendrycks, Dan and Zou, Andy and Kolter, Zico and Fredrikson, Matt and Winsor, Eric and Wynne, Jerome and Gal, Yarin and Davies, Xander},
  booktitle = {International Conference on Learning Representations (ICLR)},
  year      = {2025},
  eprint    = {2410.09024},
  archivePrefix = {arXiv},
  url       = {https://arxiv.org/abs/2410.09024}
}

@inproceedings{yuan2024rjudge,
  title     = {{R-Judge}: Benchmarking Safety Risk Awareness for {LLM} Agents},
  author    = {Yuan, Tongxin and He, Zhiwei and Dong, Lingzhong and Wang, Yiming and Zhao, Ruijie and Xia, Tian and Xu, Lizhen and Zhou, Binglin and Li, Fangqi and Zhang, Zhuosheng and Wang, Rui and Liu, Gongshen},
  booktitle = {Findings of the Association for Computational Linguistics: EMNLP 2024},
  year      = {2024},
  eprint    = {2401.10019},
  archivePrefix = {arXiv},
  url       = {https://aclanthology.org/2024.findings-emnlp.79/}
}

@inproceedings{zhang2025asb,
  title     = {Agent Security Bench ({ASB}): Formalizing and Benchmarking Attacks and Defenses in {LLM}-based Agents},
  author    = {Zhang, Hanrong and Huang, Jingyuan and Mei, Kai and Yao, Yifei and Wang, Zhenting and Zhan, Chenlu and Wang, Hongwei and Zhang, Yongfeng},
  booktitle = {International Conference on Learning Representations (ICLR)},
  year      = {2025},
  eprint    = {2410.02644},
  archivePrefix = {arXiv},
  url       = {https://arxiv.org/abs/2410.02644}
}

@inproceedings{zhan2024injecagent,
  title     = {{InjecAgent}: Benchmarking Indirect Prompt Injections in Tool-Integrated Large Language Model Agents},
  author    = {Zhan, Qiusi and Liang, Zhixiang and Ying, Zifan and Kang, Daniel},
  booktitle = {Findings of the Association for Computational Linguistics: ACL 2024},
  year      = {2024},
  eprint    = {2403.02691},
  archivePrefix = {arXiv},
  url       = {https://aclanthology.org/2024.findings-acl.624/}
}

@article{askell2021hhh,
  title     = {A General Language Assistant as a Laboratory for Alignment},
  author    = {Askell, Amanda and Bai, Yuntao and Chen, Anna and Drain, Dawn and Ganguli, Deep and Henighan, Tom and Jones, Andy and Joseph, Nicholas and Mann, Ben and DasSarma, Nova and others},
  journal   = {arXiv preprint arXiv:2112.00861},
  year      = {2021},
  url       = {https://arxiv.org/abs/2112.00861}
}

@article{bai2022constitutional,
  title     = {Constitutional {AI}: Harmlessness from {AI} Feedback},
  author    = {Bai, Yuntao and Kadavath, Saurav and Kundu, Sandipan and Askell, Amanda and Kernion, Jackson and Jones, Andy and Chen, Anna and Goldie, Anna and Mirhoseini, Azalia and McKinnon, Cameron and others},
  journal   = {arXiv preprint arXiv:2212.08073},
  year      = {2022},
  url       = {https://arxiv.org/abs/2212.08073}
}

@article{bai2022rlhf,
  title     = {Training a Helpful and Harmless Assistant with Reinforcement Learning from Human Feedback},
  author    = {Bai, Yuntao and Jones, Andy and Ndousse, Kamal and Askell, Amanda and Chen, Anna and DasSarma, Nova and Drain, Dawn and Fort, Stanislav and Ganguli, Deep and Henighan, Tom and others},
  journal   = {arXiv preprint arXiv:2204.05862},
  year      = {2022},
  url       = {https://arxiv.org/abs/2204.05862}
}

@inproceedings{rame2023rewardedsoups,
  title     = {Rewarded Soups: Towards Pareto-Optimal Alignment by Interpolating Weights Fine-Tuned on Diverse Rewards},
  author    = {Ram{\'e}, Alexandre and Couairon, Guillaume and Shukor, Mustafa and Dancette, Corentin and Gaya, Jean-Baptiste and Soulier, Laure and Cord, Matthieu},
  booktitle = {Advances in Neural Information Processing Systems (NeurIPS)},
  year      = {2023},
  eprint    = {2306.04488},
  archivePrefix = {arXiv},
  url       = {https://arxiv.org/abs/2306.04488}
}

@inproceedings{achiam2017cpo,
  title     = {Constrained Policy Optimization},
  author    = {Achiam, Joshua and Held, David and Tamar, Aviv and Abbeel, Pieter},
  booktitle = {Proceedings of the 34th International Conference on Machine Learning (ICML)},
  series    = {Proceedings of Machine Learning Research},
  volume    = {70},
  pages     = {22--31},
  year      = {2017},
  url       = {https://proceedings.mlr.press/v70/achiam17a.html}
}

@inproceedings{dai2023safe,
  title     = {Safe {RLHF}: Safe Reinforcement Learning from Human Feedback},
  author    = {Dai, Josef and Pan, Xuehai and Sun, Ruiyang and Ji, Jiaming and Xu, Xinbo and Liu, Mickel and Wang, Yizhou and Yang, Yaodong},
  booktitle = {International Conference on Learning Representations (ICLR)},
  year      = {2024},
  eprint    = {2310.12773},
  archivePrefix = {arXiv},
  url       = {https://arxiv.org/abs/2310.12773}
}

@inproceedings{zhou2024beyond,
  title     = {Beyond One-Preference-Fits-All Alignment: Multi-Objective Direct Preference Optimization},
  author    = {Zhou, Zhanhui and Liu, Jie and Shao, Jing and Yue, Xiangyu and Yang, Chao and Ouyang, Wanli and Qiao, Yu},
  booktitle = {Findings of the Association for Computational Linguistics: ACL 2024},
  pages     = {10586--10613},
  year      = {2024},
  eprint    = {2310.03708},
  archivePrefix = {arXiv},
  url       = {https://aclanthology.org/2024.findings-acl.630/}
}

@inproceedings{zhang2026darwin,
  title={{Darwin G{\"o}del Machine}: Open-Ended Evolution of Self-Improving Agents},
  author={Zhang, Jenny and Hu, Shengran and Lu, Cong and Lange, Robert Tjarko and Clune, Jeff},
  booktitle={International Conference on Learning Representations (ICLR)},
  year={2026},
  eprint={2505.22954},
  archivePrefix={arXiv},
  url={https://arxiv.org/abs/2505.22954}
}

@inproceedings{peng2025cweval,
  title={{CWEval}: Outcome-driven Evaluation on Functionality and Security of {LLM} Code Generation},
  author={Peng, Jinjun and Cui, Leyi and Huang, Kele and Yang, Junfeng and Ray, Baishakhi},
  booktitle={2025 IEEE/ACM International Workshop on Large Language Models for Code (LLM4Code)},
  pages={33--40},
  year={2025},
  organization={IEEE},
  doi={10.1109/LLM4Code66737.2025.00009},
  eprint={2501.08200},
  archivePrefix={arXiv}
}

@inproceedings{vero2025baxbench,
  title={{BaxBench}: Can {LLM}s Generate Correct and Secure Backends?},
  author={Vero, Mark and M{\"u}ndler, Niels and Chibotaru, Victor and Raychev, Veselin and Baader, Maximilian and Jovanovi{\'c}, Nikola and He, Jingxuan and Vechev, Martin},
  booktitle={Proceedings of the 42nd International Conference on Machine Learning (ICML)},
  year={2025},
  eprint={2502.11844},
  archivePrefix={arXiv},
  url={https://arxiv.org/abs/2502.11844}
}

@article{dilgren2025secrepobench,
  title={{SecRepoBench}: Benchmarking {LLMs} for Secure Code Generation in Real-World Repositories},
  author={Dilgren, Connor and Chiniya, Purva and Griffith, Luke and Ding, Yu and Chen, Yizheng},
  journal={arXiv preprint arXiv:2504.21205},
  year={2025},
  url={https://arxiv.org/abs/2504.21205}
}

@article{lee2026meta,
  title={{Meta-Harness}: End-to-End Optimization of Model Harnesses},
  author={Lee, Yoonho and Nair, Roshen and Zhang, Qizheng and Lee, Kangwook and Khattab, Omar and Finn, Chelsea},
  journal={arXiv preprint arXiv:2603.28052},
  year={2026},
  url={https://arxiv.org/abs/2603.28052}
}

@inproceedings{zhuge2024gptswarm,
  title     = {{GPTSwarm}: Language Agents as Optimizable Graphs},
  author    = {Zhuge, Mingchen and Wang, Wenyi and Kirsch, Louis and Faccio, Francesco and Khizbullin, Dmitrii and Schmidhuber, J{\"u}rgen},
  booktitle = {Proceedings of the 41st International Conference on Machine Learning (ICML)},
  series    = {Proceedings of Machine Learning Research},
  volume    = {235},
  pages     = {62743--62767},
  year      = {2024},
  eprint    = {2402.16823},
  archivePrefix = {arXiv}
}

@article{ou2025symbolic,
  title={Symbolic learning enables self-evolving agents},
  author={Ou, Yixin and Zhou, Wangchunshu and Ding, Shengwei and Li, Long and Wu, Jialong and Wang, Tiannan and Chen, Jiamin and Wang, Shuai and Xu, Xiaohua and Zhang, Ningyu and others},
  journal={AI Open},
  year={2025},
  publisher={Elsevier}
}

@article{qiu2025alita,
  title   = {{Alita}: Generalist Agent Enabling Scalable Agentic Reasoning with Minimal Predefinition and Maximal Self-Evolution},
  author  = {Qiu, Jiahao and Qi, Xuan and Zhang, Tongcheng and Juan, Xinzhe and Guo, Jiacheng and Lu, Yifu and Wang, Yimin and Yao, Zixin and Ren, Qihan and Jiang, Xun and Zhou, Xing and Liu, Dongrui and Yang, Ling and Wu, Yue and Huang, Kaixuan and Liu, Shilong and Wang, Hongru and Wang, Mengdi},
  journal = {arXiv preprint arXiv:2505.20286},
  year    = {2025},
  url     = {https://arxiv.org/abs/2505.20286}
}

@inproceedings{yin2025godel,
  title     = {G{\"o}del Agent: A Self-Referential Agent Framework for Recursively Self-Improvement},
  author    = {Yin, Xunjian and Wang, Xinyi and Pan, Liangming and Lin, Li and Wan, Xiaojun and Wang, William Yang},
  booktitle = {Proceedings of the 63rd Annual Meeting of the Association for Computational Linguistics (ACL)},
  year      = {2025},
  eprint    = {2410.04444},
  archivePrefix = {arXiv},
  url       = {https://aclanthology.org/2025.acl-long.1354/}
}

@article{robeyns2025sica,
  title   = {A Self-Improving Coding Agent},
  author  = {Robeyns, Maxime and Szummer, Martin and Aitchison, Laurence},
  journal = {arXiv preprint arXiv:2504.15228},
  year    = {2025},
  note    = {Oral, ICLR 2025 Workshop on Self-Improving Foundation Models Without Human Supervision},
  url     = {https://arxiv.org/abs/2504.15228}
}

@article{zhang2026selfharness,
  title   = {Self-Harness: Harnesses That Improve Themselves},
  author  = {Zhang, Hangfan and Zhang, Shao and Li, Kangcong and Zhang, Chen and Chen, Yang and Zhang, Yiqun and Bai, Lei and Hu, Shuyue},
  journal = {arXiv preprint arXiv:2606.09498},
  year    = {2026},
  url     = {https://arxiv.org/abs/2606.09498}
}

\appendix
\section{Notation}
\label{app:notation}

Table~\ref{tab:notation} summarizes the notation used throughout the paper.

\section{More Equations of the Problem Formulation.}
\label{append:formulations}
where
\begin{equation}
    f_{\mathrm{func}}(H)
    =
    \frac{1}{M}\sum_{n=1}^{M}
    \mathbb{E}_{\tau\sim P_H(\cdot\mid x_n)}
    \left[F(\tau,x_n)\right],
\end{equation}
and
\begin{equation}
    f_{\mathrm{sec}}(H)
    =
    \frac{1}{M}\sum_{n=1}^{M}
    \mathbb{E}_{\tau\sim P_H(\cdot\mid x_n)}
    \left[S(\tau,p_n)\right].
\end{equation}
Here $F(\tau,x_n)\in\{0,1\}$ indicates whether the solution produced in
$\tau$ passes the functional tests of $x_n$, and $S(\tau,p_n)\in\{0,1\}$
indicates whether it neutralizes the proof-of-concept exploit specified by
$p_n$. We write $f_1=f_{\mathrm{func}}$ and $f_2=f_{\mathrm{sec}}$ when
indexing objectives by $j\in\{1,\ldots,m\}$, with $m=2$; both objectives are
maximized.
In practice, we estimate these objectives using $K$ sampled trajectories
$\{\tau_{n,k}\}_{k=1}^{K}$ for each task:
\begin{equation}
    \hat f_{\mathrm{func}}(H)
    =
    \frac{1}{MK}
    \sum_{n=1}^{M}\sum_{k=1}^{K}
    F(\tau_{n,k},x_n),
    \qquad
    \hat f_{\mathrm{sec}}(H)
    =
    \frac{1}{MK}
    \sum_{n=1}^{M}\sum_{k=1}^{K}
    S(\tau_{n,k},p_n).
\end{equation}
Both estimates are pass rates in $[0,1]$, so the two objectives are on the same
scale and need no further normalization before scalarization (\S\ref{sec:moead}).

For two harnesses $H,H'\in\mathcal{H}$, $H$ \textbf{Pareto-dominates} $H'$, written
$H\succ H'$, if
$f_j(H)\geq f_j(H')$ for every objective $j$, with strict inequality for at
least one. The Pareto set is
\begin{equation}
\label{eq:Pareto-dominates}
    \mathcal{P}^{*}
    =
    \left\{
        H\in\mathcal{H}
        \mid
        \nexists H'\in\mathcal{H}: H' \succ H
    \right\},
\end{equation}
and its image $\{\mathbf{f}(H):H\in\mathcal{P}^{*}\}$ defines the Pareto
front. Our goal is to evolve a finite set of harnesses that provides a
high-quality approximation to this front.

\begin{table}[!htbp]
\centering
\small
\caption{Summary of notation.}
\label{tab:notation}
\renewcommand{\arraystretch}{1.12}
\begin{tabularx}{\linewidth}{@{}lL@{}}
\toprule
\textbf{Symbol} & \textbf{Meaning} \\
\midrule
\multicolumn{2}{@{}l}{\emph{Problem formulation (\S\ref{sec:formulation})}} \\
$\mathcal{H}$, $H$ & Space of agent harnesses; a harness \\
$\mathcal{D}=\{t_n\}_{n=1}^{M}$ & Evaluation task set of $M$ tasks \\
$t_n=(x_n,p_n)$ & Task $n$: task specification $x_n$ and security policy $p_n$ \\
$\tau\sim P_H(\tau\mid x_n)$ & Trajectory of harness $H$ on task $t_n$ (conditioned on the visible spec $x_n$) \\
$F(\tau,x_n)\in\{0,1\}$ & Functional success: the solution passes the tests of $x_n$ \\
$S(\tau,p_n)\in\{0,1\}$ & Security success: the solution neutralizes the exploit of $p_n$ \\
$\mathbf{f}(H)=(f_{\mathrm{func}},f_{\mathrm{sec}})$ & Expected functionality and security; $f_1=f_{\mathrm{func}}$, $f_2=f_{\mathrm{sec}}$ \\
$\hat f_j$, $K$, $\tau_{n,k}$ & Monte Carlo estimate of $f_j$ from $K$ sampled trajectories $\tau_{n,k}$ per task \\
$m$, $j$ & Number of objectives ($m=2$); objective index \\
$H\succ H'$, $\mathcal{P}^{*}$ & Pareto dominance; Pareto set \\
$g$, $\mathcal{U}_g$, $\mathcal{U}$ & Generation index; children evaluated in generation $g$; all evaluated children \\
$H^{\dagger}$ & Deployed harness: most balanced child in $\mathcal{U}$ on the search split (Eq.~\ref{eq:selection}) \\
\midrule
\multicolumn{2}{@{}l}{\emph{Decomposition-based evolution (\S\ref{sec:what}--\S\ref{sec:moead})}} \\
$\mathcal{K}$ & Component classes $\{\text{tools},\text{skills},\text{logic}\}$ (logic includes orchestration) \\
$H_0$ & Initial (seed) harness \\
$\Lambda=\{\boldsymbol{\lambda}^{1},\ldots,\boldsymbol{\lambda}^{N}\}$ & $N$ weight vectors on the $m$-objective simplex, one per subproblem \\
$i$, $\lambda^{i}_{j}$ & Subproblem index; weight of objective $j$ in subproblem $i$ \\
$H_i$, $H_i^{*}$ & Incumbent harness of subproblem $i$; its (conceptual) optimum \\
$\mathbf{z}^{*}$, $z^{*}_{j}$ & Ideal point; best observed value of objective $j$ \\
$g^{\mathrm{te}}(H\mid\boldsymbol{\lambda}^{i},\mathbf{z}^{*})$ & Weighted Tchebycheff scalarization of subproblem $i$ (lower is better) \\
$B(i)$, $T$, $l$ & Neighborhood of subproblem $i$; its size; neighbor index $l\in B(i)$ \\
$H_c$ & Child harness produced by variation \\
\midrule
\multicolumn{2}{@{}l}{\emph{Factorized evolution (\S\ref{sec:operators})}} \\
$\mathcal{E}_i$ & Execution feedback collected from the parent of subproblem $i$ \\
$\mathcal{Q}$, $d_i$ & Diagnosis step; failure diagnosis $d_i=\mathcal{Q}(\mathcal{E}_i)$ \\
$\mathcal{G}$, $h_i$ & Hypothesis step; improvement hypothesis $h_i=\mathcal{G}(d_i,\boldsymbol{\lambda}^{i})$ \\
$\pi_{\mathrm{comp}}$, $k_i\in\mathcal{K}$ & Component selection; selected component class \\
$\Sigma=\{\mathrm{gen},\mathrm{env}\}$ & Adaptation scopes: general and environment-specific \\
$\mathcal{R}_{k_i}$, $s_i\in\Sigma$ & Component-specific scope assignment; selected scope \\
$\mathcal{O}_{k,s}$ & Variation operator editing component class $k$ within scope $s$ \\
$\mathcal{O}_{\mathrm{generalize}}$ & Cross-scope operator moving components between $\mathrm{gen}$ and $\mathrm{env}$ \\
$\Pi=(\mathcal{O}_1,\ldots,\mathcal{O}_L)$ & Ordered operator plan of length $L$ \\
$H_c^{(\ell)}$ & Candidate after the $\ell$-th operator; $H_c^{(0)}=H_i$, $H_c=H_c^{(L)}$ \\
\midrule
\multicolumn{2}{@{}l}{\emph{Dynamic composition (\S\ref{sec:composition})}} \\
$e$, $x$ & Execution environment (benchmark); task within it \\
$\mathcal{H}^{\mathrm{gen}}$, $\mathcal{H}^{\mathrm{env}}_{e}$ & Pools of general and environment-specific components \\
$\mathcal{H}^{\mathrm{gen}}_{k}$, $\mathcal{H}^{\mathrm{env}}_{k,e}$ & Component pools restricted to class $k$ \\
$\pi_{\phi}$, $\mathcal{A}_{e,x}$ & Composition policy; components activated for $(e,x)$ \\
$\mathcal{C}$, $H^{(e,x)}$ & Runtime composition procedure; executable harness for $(e,x)$ \\
\midrule
\multicolumn{2}{@{}l}{\emph{Evaluation and overfitting gate (\S\ref{sec:experiments}, App.~\ref{sec:appendix-plants})}} \\
$F\wedge S$ & Secure-pass@1: functional and secure on the same task \\
$\Delta_{\mathrm{evolve}}$, $\Delta_{\mathrm{OUT}}$ & Security ($f_{\mathrm{sec}}$) gain over baseline on the evolve split and on held-out OUT (Table~\ref{tab:retention}) \\
$\delta$, $\rho(\delta)$, $\theta$ & Parent-to-child diff; judged overfitting risk; rejection threshold \\
$C$, $C^{+}_{q}$, $n_C$ & Admitted child; its copy with injected overfit of type $q$; number of children \\
$s_{\mathrm{evolve}}$, $s_{\mathrm{ho}}$ & Secure-pass@1 on the evolve split and on held-out IN$\,\cup\,$OUT \\
$\Delta^{\wedge}_{\mathrm{evolve}}$, $\Delta^{\wedge}_{\mathrm{ho}}$, $\Delta_{\mathrm{gap}}$ & Secure-pass@1 gain of $C^{+}$ over $C$ on the evolve and held-out splits; generalization gap $\Delta^{\wedge}_{\mathrm{evolve}}-\Delta^{\wedge}_{\mathrm{ho}}$ \\
\midrule
\multicolumn{2}{@{}l}{\emph{Population diversity (App.~\ref{sec:diversity})}} \\
$g_{\mathrm{sh}}$ & Token $g_{\mathrm{sh}}$-gram (shingle) length used by the diversity measure (not to be confused with the generation index $g$) \\
$\epsilon_{\mathrm{dup}}$ & Near-duplicate distance threshold defining equivalence classes \\
\bottomrule
\end{tabularx}
\end{table}

\clearpage
\section{Implementation Details}

\textbf{Benchmarks and Setup.}
\label{sec:exp-setup}
We conduct our main experiments on two complementary secure-code benchmarks:
\textbf{SecRepoBench}~\citep{dilgren2025secrepobench} and
\textbf{BaxBench}~\citep{vero2025baxbench}.
SecRepoBench evaluates repository-level secure code completion in real-world
C/C++ projects: given a masked code region and repository context, an agent
must produce a patch that preserves functional correctness while eliminating a
known vulnerability, with both properties verified through execution-based unit
tests and proof-of-concept exploits. BaxBench instead evaluates the generation
of complete backend applications, testing functional behavior with end-to-end
test suites and security against executable attacks. Together, the two
benchmarks cover substantially different settings---repository-level program
repair and application-level generation---while exposing the same underlying
functionality--security trade-off. To assess transfer beyond the environments, we additionally
evaluate evolved harnesses on \textbf{CWEval}~\citep{peng2025cweval}, an
outcome-driven secure code-generation benchmark that jointly evaluates
functional correctness and vulnerability-specific security using executable
tests.

\paragraph{Baselines.} Meta-harness~\citep{lee2026meta} and
DGM~\citep{zhang2026darwin} are not reimplementations: both run from their
authors' released code, with our benchmarks and grader plugged in as the
evaluation.

\paragraph{Repeats and effective denominators.} Every reported rate is the mean
of \textbf{three independent runs}. Grading depends on live infrastructure---a
task is scored only if its ARVO image pulls and its build and exploit probe
complete---and a small number of tasks intermittently fail for reasons unrelated
to the harness under test (registry pull failures, and timeouts caused by
contention on the local CPU during heavy rebuilds such as \texttt{ffmpeg}). Those
tasks are excluded from the run in which they failed, so the effective
denominator of a cell is at most, and occasionally below, the $30$ (SecRepoBench)
or $28$ (BaxBench) tasks in its split; this is why some rates are not integer
multiples of $1/30$ or $1/28$. Infrastructure failures are independent of the
method being evaluated, but they are not independent \emph{across} methods.

Every row in the lower block of Table~\ref{tab:strategy} shares the full configuration's factored evolving (\S\ref{sec:operators}) and dynamic orchestration (\S\ref{sec:composition}). 

\section{Details for Decomposition}

A linear scalarization,
$\sum_j \lambda^{i}_{j} f_j(H)$, can recover only supported Pareto-optimal solutions
and may therefore miss entire Pareto-optimal regions when the frontier is
non-convex. In contrast, the weighted Tchebycheff formulation does not rely on
convexity of the Pareto frontier and can recover solutions in such regions.
This guarantee matters only when the objectives actually trade off. On our
benchmarks the observed fronts are narrow: on the SecRepoBench search split the
front holds two non-dominated harnesses in early generations, and in the last
generation one harness is best on both objectives (Table~\ref{tab:pareto},
\S\ref{sec:exp-pareto}). What
decomposition still provides there is a search direction: each subproblem
measures progress by its worst weighted shortfall from $\mathbf{z}^{*}$, so a
child that improves one objective by giving up the other is not accepted.
Moreover, each $\boldsymbol{\lambda}^{i}$ defines an independent search
direction: no single scalarization is treated as the global objective, and the
population jointly maintains solutions corresponding to different regions of
the frontier.

\section{Implementation Details for Factorized Harness Evolution}
\label{app:factorized-evolution}

This appendix provides implementation details for the factorized evolution
procedure introduced in Section~\ref{sec:operators}.

\paragraph{Harness Initialization.} \label{app:init}
We initialize evolution from a lightweight but functional base harness
$H_{0}$. The base agent follows a standard ReAct-style interaction loop and
provides only a minimal set of general-purpose tools---including file reading,
file writing, and command-line execution---together with a simple orchestrator
for sequencing reasoning, tool use, and execution. This deliberately modest
initialization provides the basic interface required to solve tasks while
leaving the agent's higher-level tools, control logic, verification mechanisms,
and workflow structure to be discovered through evolution. The initial
incumbents of the $N$ decomposition subproblems are instantiated from this base
harness.

\paragraph{Harness architecture.}
The harness is a self-contained Python package.
The package itself is the evolvable genome: the evaluator imports only the
\texttt{AgentHarness} entry point, and nothing outside the package is imported,
so every part of the solver can be changed by the evolutionary operators. That
includes the LLM call layer, the tools, the prompts and the control logic
(Figure~\ref{fig:harness-layout}). 

\begin{figure}[th]
\centering
\small
\begin{verbatim}
Harness/          # the genome: one self-contained package
|-- config/           # frozen agent config + per-benchmark overrides
|-- orchestrator/     # LLM planner, DAG validation, wave executor
|-- steps/            # registry of composable workflow nodes
|-- tools/            # real tools: docker build, bash, read, edit
|-- skills/           # injectable domain expertise
|-- prompts/          # system / generation / repair / audit templates
`-- utils/            # stateless helpers (parsing, messages, trajectory)
    (each directory = general/ + <benchmark>/)
\end{verbatim}
\caption{Top-level layout of the harness. The entry point
(\texttt{AgentHarness}) and the LLM call layer sit at the package root.}
\label{fig:harness-layout}
\end{figure}

\subsection{Diagnose Execution Feedback}
\label{app:proposer-diagnosis}

Given the failure diagnosis, the proposer selects one or more
component-specific operators and the scope in which each operator should act.
The \texttt{tool\_evolve} operator modifies executable tools and their
integration with the agent loop. The \texttt{skill\_evolve} operator modifies
reusable procedures, domain knowledge, and skill modules. The
\texttt{logic\_evolve} operator modifies prompting, reasoning procedures,
verification logic, control flow, and orchestration. Each operator is restricted
to its corresponding portion of the harness rather than being instructed to
rewrite the repository as a single artifact.

Before \emph{every} variation step we run a diagnosis on the niche incumbent
(\texttt{diagnose.py}). A proposer session with \emph{read-only} tools
(\texttt{Read}/\texttt{Glob}/\texttt{Grep}, sandboxed away from other runs' trees and from the
harness's own notes) is given the subproblem's weight vector $\boldsymbol{\lambda}^{i}$, the artifact-grounded failure
digest, and the incumbent's genome: inlined verbatim when it is
small ($\le 8$K characters) and otherwise replaced by a file manifest plus the
\emph{force-inlined} entry-point modules ($\le 24$K), since an instruction to read the files
is empirically ignored while inlined text cannot be skipped. Three hard constraints are then
stated immediately before the reply schema. (i) A \emph{niche lens} fixes which objective this
subproblem may diagnose: the weight vector decides, and only for the exactly balanced
subproblem is the tie broken by the parent's own worst objective; the same lens builds the
digest, so ``what counts as a failure'' and ``what may be cited'' cannot drift apart.
(ii) The diagnoses already issued by sibling subproblems \emph{in this generation} are shown
with the order to pick a different lever, because otherwise one generation produces one child
in five copies. (iii) When the executing operator is fixed and hard-gated, its delivery
contract is inlined (for \texttt{tool}, together with the measured call-site/implementation
inventory) and the \texttt{focus} enum is narrowed to what that operator can actually deliver
--- an undeliverable plan costs the niche without the operator ever having had a legal move.
The reply must be a single JSON object under 6K characters:
\[
\begin{aligned}
\{\,&\texttt{focus}\in\mathcal{F}_{\mathrm{focus}},\ \texttt{problem\_statement},\
\texttt{artifact\_evidence},\\
&\texttt{few\_shot\_examples\_of\_failure}\ [,\ \texttt{operators}]\,\},
\end{aligned}
\]
$\mathcal{F}_{\mathrm{focus}}=\{$tools, reliability, verification, prompt, control-flow, skills, failure-fix$\}$,
where \texttt{few\_shot\_examples\_of\_failure} is $1$--$3$ concrete failures as
$\{$task\_id, symptom, direct\_cause, evidence$\}$ --- category-level non-causes (``the model
made a mistake'', ``insufficient validation'') are explicitly rejected --- and
\texttt{operators} (only under \texttt{--operator auto}) is an ordered pipeline of up to
$8$ steps $\{$op, goal$\}$, each step editing the previous step's output with a distinct
sub-goal and only the final result evaluated.

Two mechanisms make the plan trustworthy rather than merely well-formed. The reply is parsed
by scanning \emph{backwards} for the last JSON object that decodes and carries a schema key
(brace-balance tracking was removed: one unmatched brace quoted from source code in the
narration poisoned every later candidate), with a repair path for replies whose opening brace
was truncated and alias normalization for synonymous field names; ``parsed'' and ``usable''
(can it actually steer the operator) are recorded as separate columns. On top of that, an
\emph{evidence-integrity gate} validates \texttt{artifact\_evidence} purely on the filesystem:
the cited \texttt{path} must exist under this parent's own eval-artifact tree, and the
\texttt{quote} ($\ge 24$ characters) must occur verbatim in it after whitespace normalization.
A reply that is unusable, oversized, empty, or that cites unverifiable evidence triggers
\emph{one} retry whose correction names that specific failure; a chain that came back in an
invented schema triggers a second, chain-only re-ask. If the evidence still does not check
out, the plan's guidance is \emph{suppressed} and the operator proceeds unguided rather than
being handed unproven fact --- the same end state as an unusable reply, and recorded as such.
Everything the diagnosis touches is pure guidance: it never enters MOEA/D selection, every
failure path degrades to the empty block, and each call leaves a readable scene in the niche's
own directory (\texttt{diagnose\_input.md} / \texttt{diagnose\_output.md} / \texttt{chain.md})
plus one line in \texttt{logs/moead\_diagnose.jsonl}, so lens, focus duplication, parse/usable
rate, gate firing rate and retry counts are all countable per run.

\subsection{Tool Evolution, Skill Evolution and Logic Evolution}
\label{app:tool-evolve}

Usually multiple operators are selected; they are applied sequentially, with each
operator receiving the latest candidate produced by the preceding operator,
and $H_c$ is the output of the last one.
This permits coordinated changes across multiple components while keeping each
individual edit semantically scoped.

The decomposed operators are the three \emph{scoped} variation operators
summarized in Table~\ref{tab:scoped-operators}: \texttt{tool}, \texttt{skill},
and \texttt{logic}. Each one targets a distinct part of the harness and may
write only to the directories assigned to it. Before any edit, the operator
must also decide whether the change belongs to the general scope
(\texttt{general/}) or to the benchmark-specific scope
(\texttt{\textless bench\textgreater/}).

\paragraph{Scope selection.}
The scope is fixed before editing by \texttt{decide\_scope}, a short, read-only
model call that answers a single question: \emph{can the agent derive the
relevant fact at run time on its own?} If it can, the change is placed in the
general scope. If the fact is instead an intrinsic rule of the evaluator or of
the delivery channel, the change is placed in the benchmark-specific scope.
The chosen scope is then compiled into a write whitelist, and the whitelist is
enforced at the tool layer, not through the prompt: any file written outside
it is reverted. If the \texttt{decide\_scope} call fails, the operator falls
back to the general scope, which is the more restrictive of the two.

The two scopes impose different obligations on the edit:
\begin{itemize}[leftmargin=1.4em,itemsep=2pt,topsep=2pt]
  \item \textbf{General scope.} The operator is instructed not to hard-code any
        fact particular to the current benchmark, such as port numbers,
        framework names, or CWE lists. Any such fact the change requires must
        be obtained at run time.
  \item \textbf{Benchmark-specific scope.} The edit may be as specific as the
        benchmark demands, but each file must state in its header which
        benchmark it targets. A module in this scope shadows the general module
        of the same name, so benchmark-specific behavior overrides the general
        default without modifying it.
\end{itemize}

\begin{table}[t]
\centering
\small
\caption{The three scoped variation operators. Each operator may write only its own
directories, at the scope chosen before the edit; any write outside them is reverted.}
\label{tab:scoped-operators}
\renewcommand{\arraystretch}{1.15}
\begin{tabularx}{\linewidth}{@{}l p{3.3cm} L@{}}
\toprule
\textbf{Operator} & \textbf{Writable directories} & \textbf{Required deliverable} \\
\midrule
\texttt{tool}
  & \texttt{tools/}, \texttt{utils/} (plus \texttt{config/})
  & A change to how the agent \emph{obtains or consumes ground truth}, covering one
    of four cases: no real tool is wired into the loop; a tool is called but its
    result is wasted (the log is ignored, or only the build is checked and never the
    server's startup); the tool itself returns too little to act on; or a helper in
    \texttt{utils/} discards the critical information before it reaches the agent.
    Phantom verification is explicitly forbidden. \\
\midrule
\texttt{skill}
  & \texttt{skills/}, \texttt{prompts/}
  & \emph{One} reusable skill: a trigger condition, a short general procedure, and
    one worked example. The skill is a single \texttt{.py} module dropped into the
    scope directory, with no registration needed. Skills written as a
    \emph{default behavior with an escape clause} (e.g.\ ``unless the spec says
    otherwise, return 200'') are forbidden. In one of our runs, such a skill fixed its target
    task but broke two unrelated ones. \\
\midrule
\texttt{logic}
  & \texttt{steps/}, \texttt{orchestrator/}; at \texttt{general} scope, also the
    loop modules at the package root
  & A change to the \emph{control flow}: add a missing stage, act on a branch that
    is currently ignored, adjust round counts and budgets (the values go in
    \texttt{config/}), or change the DAG itself. Achieving the same effect by
    editing prompts is not allowed. \\
\bottomrule
\end{tabularx}
\end{table}

\subsection{Generalization Operator}
\label{app:generalize}

The operator evolves for dependencies on incidental properties of the
evolution environment, such as hard-coded repository paths, module names,
benchmark-specific commands, magic strings, task-specific instructions,
over-tuned thresholds, or parsers that assume a particular repository
structure or test runner. When such behavior is useful only in one environment,
it can be localized to the corresponding environment-specific scope. When the
underlying mechanism can be expressed independently of that environment, the
operator may abstract or parameterize it and move the resulting mechanism into
the general scope. Conversely, a component previously placed in the general
scope may be localized if evolution reveals that it relies on
benchmark-specific assumptions.

Thus, \texttt{generalize} does not attempt to recover a theoretically defined
partition between general and specialized behavior. Instead, it makes the
boundary itself subject to evolutionary reorganization. 

The scoped operators each add capability to the harness. \texttt{generalize}
adds none. It changes only the scope in which existing components are
\emph{filed}. Each invocation audits both partitions of the harness, the
general scope and the benchmark-specific scope, and corrects exactly \emph{one}
misfiled unit. Which move applies depends on two properties of the unit: the
scope in which it currently resides, and whether the benchmark-specific
knowledge it encodes can be recovered at run time.

\begin{itemize}[leftmargin=1.4em,itemsep=2pt,topsep=2pt]
  \item \textbf{Promote} (benchmark-specific $\rightarrow$ general).
        \emph{Condition:} a procedure that is general in nature is filed under
        the benchmark-specific scope, so agents deployed on other benchmarks
        cannot use it. \emph{Correction:} the unit is moved to the general
        scope. Each benchmark-specific fact it depends on is replaced by a step
        that obtains that fact at run time, so the promoted unit no longer
        assumes the benchmark it was written for.
  \item \textbf{Demote} (general $\rightarrow$ benchmark-specific).
        \emph{Condition:} a unit in the general scope hard-codes a fact that
        holds only on the current benchmark, and the fact cannot be recovered
        at run time. \emph{Correction:} the fact is moved to the
        benchmark-specific scope, where a tunable value becomes a per-benchmark
        configuration override. The general unit then reads the value from
        that scope instead of stating it.
  \item \textbf{Derive} (remains general).
        \emph{Condition:} a unit in the general scope hard-codes a
        benchmark-specific fact that \emph{can} be recovered at run time.
        \emph{Correction:} the unit stays in the general scope and is rewritten
        to obtain the fact at run time, which removes the dependency without
        adding any benchmark-specific state.
\end{itemize}

\noindent
Demote and Derive thus address the same symptom, a general unit that encodes
benchmark-specific knowledge, and are distinguished only by whether that
knowledge is observable at run time. When it is, Derive is preferred because it
leaves nothing in the benchmark-specific scope.

Every move is subject to two invariants. \emph{(i) Behavioral equivalence on
the current benchmark.} The benchmark-specific partition is still loaded on
the current benchmark, so a correct reclassification leaves the search-split
score unchanged. Its benefit is observable only on held-out tasks and when the
general partition is evaluated in isolation. \emph{(ii) Standalone
executability of the general partition.} Loading the general scope alone must
still yield a functioning agent, even if it performs worse.

\subsection{Operator Scheduling}
\label{app:operator-scheduling}

\paragraph{Operator scheduling.}
The component-level operators in \methodname are applied sequentially rather
than simultaneously. Given the incumbent $H_i$ of subproblem $i$, objective
preference $\boldsymbol{\lambda}^{i}$, and execution feedback $\mathcal{E}_i$, the
proposer first produces an ordered operator plan
\begin{equation}
    \Pi
    =
    \left(
        \mathcal{O}_1,\ldots,\mathcal{O}_L
    \right),
    \qquad
    \mathcal{O}_{\ell}
    \in
    \{
        \mathcal{O}_{k,s}
        :
        k\in\mathcal{K},\,
        s\in\Sigma
    \}
    \cup
    \{
        \mathcal{O}_{\mathrm{generalize}}
    \}.
\end{equation}
Both the subset of operators and their ordering are determined by the proposer
from its diagnosis of the current failures. Starting from
$H_c^{(0)}=H_i$, the operators are applied sequentially,
\begin{equation}
    H_c^{(\ell)}
    =
    \mathcal{O}_{\ell}
    \left(
        H_c^{(\ell-1)};
        h_i
    \right),
    \qquad
    \ell=1,\ldots,L,
\end{equation}
where $h_i$ is the improvement hypothesis of \S\ref{sec:operators}, which already
conditions on $\boldsymbol{\lambda}^{i}$ and $\mathcal{E}_i$. The child is
$H_c=H_c^{(L)}$.
Consequently, each operator observes the modifications introduced by previous
operators, allowing later edits to reconcile or build upon earlier ones rather
than producing independent patches that must subsequently be merged.

\subsection{Validation, Repair, and Rollback}
\label{app:validation-recovery}
\paragraph{Validation and recovery.}
After executing the proposed operator sequence, the resulting child enters a
three-stage smoke-test pipeline that progressively checks the integrity of the
modified harness. If any stage fails, the resulting execution trace and error
diagnostics are returned to the proposer, which is given one opportunity to
repair the candidate. The complete three-stage validation is then repeated.
A candidate that fails again is rejected, and the working repository is restored
to the last valid parent version. Only candidates that pass all validation
stages proceed to the full task-level evaluation used by MOEA/D.

No stage calls an LLM or starts a benchmark container, so the whole pipeline
takes seconds to a few minutes per child.

\textbf{Stage 1: import and interface validation.} The child package is
imported, together with every submodule individually, because deferred imports
(e.g.\ the orchestrator and step registry) are not covered by importing the
entry point. The frozen evaluator interface is then checked: the
\texttt{AgentHarness} class must expose an asynchronous \texttt{run} (or agent
loop) method, and its user-prompt template must still contain the task
specification slot.

\textbf{Stage 2: offline smoke test and genome self-tests.} The harness is
instantiated against a stub environment and must embed a placeholder
specification into its messages and parse a canned model reply into files. The
genome's own \texttt{\_selftest*.py} files are then run as subprocesses
(timeout $300$\,s per file). The seed ships a self-test that drives the full
tool-calling loop against a scripted stub model instead of a real LLM. It checks
that the tool schemas exposed to the model are well-formed, that the payloads sent
to each vendor have the shape that vendor requires (e.g.\ every tool use paired
with a tool result), and that the loop terminates within budget and actually
modifies the app. Without this check, a broken tool-calling protocol is swallowed
by the agent's catch-all and scores a whole generation $0/0$.

\textbf{Stage 3: static integrity checks.} Two AST-only checks cover what the
first two stages cannot execute. First, every named self-test assertion of the
parent must still exist in the child. The self-tests are part of the evolvable
genome, and the cheapest way to make a failing self-test pass is to delete
it. Second, a static call-signature check requires every call in the child to bind to
the definition it resolves to, which catches half-edited call sites on branches
the smoke test never reaches.

A child that fails any stage receives one repair attempt with the failing
stage's log; a child that passes but is byte-identical to its parent is rejected
as well.

This mechanism is particularly important when multiple operators are scheduled
for the same child: sequential editing ensures that each modification is
conditioned on the current harness state, while end-to-end validation verifies
that their composition remains executable before the candidate can affect the
evolutionary population.

\section{Implementation Details for Dynamic Composition}
\label{app:dynamic-composition}

This appendix provides implementation details for the dynamic composition
procedure introduced in Section~\ref{sec:composition}.

Operationally, this happens in two steps. \emph{Scope loading} fixes the pools:
every component directory of the harness is split into a \texttt{general/}
part and one part per benchmark, and at load time the harness imports
\texttt{general/} plus the directory of the \emph{active} environment $e$ only.
\emph{Task-level planning} then implements $\pi_\phi$. An LLM planner reads the
menu of registered workflow steps together with the task specification and emits
a task-specific DAG whose nodes are those steps (e.g.\ generate, build-verify,
security audit, repair), with optional \texttt{when}-guarded conditional edges.
$\mathcal{A}_{e,x}$ is the set of steps, and the tools and skills they load, that
appear in this DAG. $\mathcal{C}$ is the executor that validates the DAG and runs
it in topological order. When the harness is transferred zero-shot to an
environment $e'$ it was not evolved on, the
source benchmark's directories are not loaded and $\mathcal{H}^{\mathrm{env}}_{e'}$
is empty, so only $\mathcal{H}^{\mathrm{gen}}$ transfers.

\subsection{Composition Agent}
\label{app:composition-agent}

We organize the evolvable harness into three component classes: tools, skills,
and logic. The logic component includes prompting, verification, control flow,
and high-level orchestration. For each class, the repository may contain both
general components and components specialized to a particular execution
environment $e$. We denote the corresponding component pools by
\begin{equation}
\label{eq:factorized-harness}
\mathcal{H}^{\mathrm{gen}}
=
\bigcup_{k\in\mathcal{K}}
\mathcal{H}_k^{\mathrm{gen}},
\qquad
\mathcal{H}^{\mathrm{env}}_{e}
=
\bigcup_{k\in\mathcal{K}}
\mathcal{H}_{k,e}^{\mathrm{env}},
\end{equation}
from which the composition agent of Eq.~\ref{eq:component-selection} selects
the active components.

\subsection{Runtime Harness Assembly}
\label{app:runtime-composition}

\paragraph{Operational factorization of the harness.}
$\mathcal{H}^{\mathrm{gen}}$ and $\mathcal{H}^{\mathrm{env}}_{e}$ are not required to be two
standalone programs or functions. Instead, they correspond to logically distinct
regions of the same harness repository. We bias executable tooling and reusable
agent-control mechanisms toward $\mathcal{H}^{\mathrm{gen}}$, since abstractions such as
execution, verification, retry, and context management should ideally transfer
across environments. In contrast, retrieved skills and benchmark-specific
procedures---including environment conventions, task-specific recipes, and
local execution knowledge---form $\mathcal{H}^{\mathrm{env}}_{e}$. Logic and orchestration
may coordinate both components and therefore need not belong exclusively to
either category.

Because this distinction is inherently semantic rather than syntactic, we do
not impose a rigid file-level partition. Instead, the
\texttt{generalize} operator serves as an explicit mechanism for maintaining
the boundary. It audits the current harness for dependencies whose utility
arises from incidental properties of the evolution environment and either
rewrites them into environment-agnostic mechanisms or confines the
specialization to environment-specific components. Thus, factorization is
enforced through evolutionary pressure and operator scope rather than through a
fixed manually specified decomposition of the repository.

\section{Additional results}
\label{sec:appendix-tables}

\begin{table}[ht]
\centering
\caption{Component ablation of joint evolution on SecRepoBench (search split, \%). Each row adds one component to the row above. The bottom row (+ dynamic orchestration) is the full method and is repeated from Table~\ref{tab:evolve-secrepo} (mean over 3 runs).}
\label{tab:main-ablation}
\small
\begin{tabular}{llccc}
\toprule
Strategy & Acceptance rule & Func & Sec & F\&S \\
\midrule
Joint Evolving              & Tchebycheff over $(F,S)$     & 70.0 & 76.7 & 63.3 \\
\quad + factored evolving       & \; + factorized operators (\S\ref{sec:operators})     & 70.0 & 83.3 & 66.7 \\
\quad + dynamic orchestration   & \; + dynamic composition (\S\ref{sec:composition}) & \textbf{76.7} & \textbf{93.3} & \textbf{76.7} \\
\bottomrule
\end{tabular}
\end{table}

\begin{table}[ht]
\centering
\caption{Operator-pool ablation (F\&S, \%). ``Gen-3 children'' lists the F\&S of each child produced in the third generation; ``Best'' is the best F\&S reached by the run. $\dagger$: run that leads to fallback.}
\label{tab:operator-pool}
\small
\begin{tabular}{clcccc}
\toprule
Orch. & Operator pool & Seed & Gen-3 children & Best \\
\midrule
\cmark & \texttt{tool}, \texttt{skill}, \texttt{logic}       & 66.7 & 83.3, 83.3, 33.3        & 83.3 \\
\cmark & \quad + \texttt{generalize}                           & 66.7 & 100.0, 50.0, $\dagger$  & \textbf{100.0} \\
\xmark & \texttt{tool}, \texttt{skill}, \texttt{logic}       & 33.3 & 66.7, 66.7, 66.7        & 66.7 \\
\xmark & \quad + \texttt{generalize}                         & 33.3 & 66.7, 83.3, 66.7  & 83.3 \\
\bottomrule
\end{tabular}
\end{table}

\paragraph{Effect of \texttt{generalize}.}
Because each run requires full container-based evaluation, we conduct this ablation on a 6-task subset of the search split, where a single task corresponds to 16.7 points. Table~\ref{tab:operator-pool} compares evolution with and without the \texttt{generalize} operator, with the orchestrator on and off; both runs within each pair start from the same seed. Adding \texttt{generalize} improves the best F\&S in both settings: from 83.3 to 100.0 with the orchestrator on, and from 66.7 to 83.3 with it off. \texttt{generalize} adds no new capability of its own: it only moves existing components between the general and environment-specific scopes (App.~\ref{app:generalize}). We therefore attribute the improvement to its effect on subsequent edits: once benchmark-specific assumptions are removed from the general scope, later operators build on mechanisms that hold across tasks rather than on fixes that happen to work for a few. Given the small subset, each gain amounts to one additional task solved, so we read these results as a consistent directional signal across both orchestrator settings rather than as a precise effect size.

\subsection{Evolving on one objective/distribution degrades safety elsewhere}
\label{sec:exp-motivation}
Before proposing joint evolution, we confirm the problem exists. A harness evolved
to maximize its \emph{search-set} score (the ``baseline-seeded winner'') collapses
on held-out data: combined F\&S falls to $35.0$, \emph{below} the un-evolved
baseline's $43.3$ (Table~\ref{tab:heldout}). Qualitatively (\S\ref{sec:casestudy}),
an agent that only ever sees functional feedback writes plausible code, declares
completion, and stops---with nothing pushing it toward the missing hardening
(e.g., a bounds check), because the exploit signal is hidden at solve time. This
is the self-evolution analogue of overfitting: capability specialized to one
distribution, security left behind.
%

\begin{figure}[t]
\centering
\includegraphics[width=0.62\linewidth]{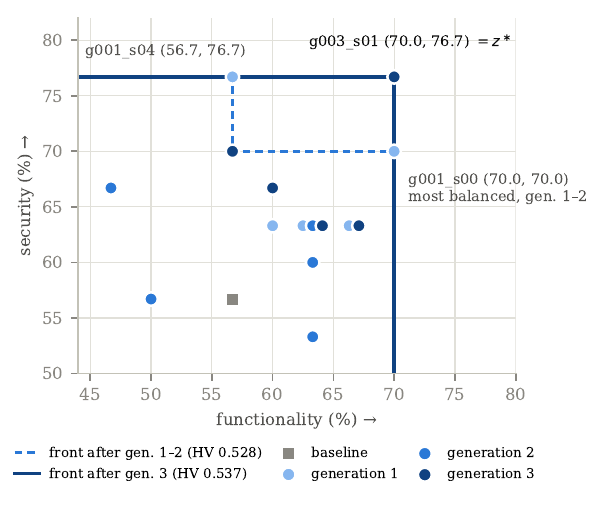}
\caption{\textbf{Search-split Pareto front of the MOEA/D-alone run}
(Table~\ref{tab:pareto}): the baseline and all $15$ children, shaded by generation.
After generations 1--2 the front holds two non-dominated harnesses (dashed); in
generation 3, \texttt{moead\_g003\_s01} reaches the ideal point $z^\ast$ and
dominates the rest (solid). Children with identical scores are offset
horizontally so that each stays visible. An interactive version is available at
\url{https://anonymous.4open.science/w/HH_visualization-4F69/index.html}.}
\label{fig:pareto-secrepo}
\end{figure}

\begin{figure}[t]
\centering
\includegraphics[width=0.62\linewidth]{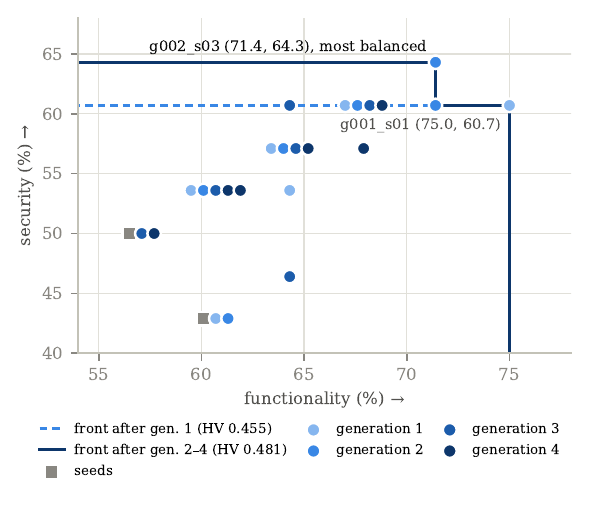}
\caption{\textbf{Search-set Pareto front of a MOEA/D run on BaxBench} ($N=6$
subproblems, 28-task cover set, generations 1--4): the two seed harnesses and all
$24$ children, shaded by generation. After generation 1 the front is a single
harness (dashed); from generation 2 on it holds two mutually non-dominated harnesses
(solid). Children with identical scores are offset horizontally so that each stays
visible. An interactive version is available at
\url{https://anonymous.4open.science/w/HH_visualization-4F69/baxbench.html}.}
\label{fig:pareto-baxbench}
\end{figure}

\textbf{Multi-objective evolution and the Pareto frontier.}
\label{sec:exp-pareto}
To isolate the decomposition, this analysis runs MOEA/D alone ($N=5$
subproblems, neighborhood $T=2$, 3 generations): Tchebycheff joint evolution
without factored evolving (\S\ref{sec:operators}) or dynamic orchestration
(\S\ref{sec:composition}), so that the frontier reflects the multi-objective
search and not the other components. Starting from the
un-evolved baseline, hypervolume jumps in generation 1 ($0.321\!\to\!0.528$) and
plateaus ($\to\!0.537$). Table~\ref{tab:pareto} lists one harness per
generation, the most balanced one under Eq.~\ref{eq:selection}, not the whole
front. The front itself is wider: in generations 1 and 2 it holds two
non-dominated harnesses, the most balanced one at $(\text{func},\text{sec}) =
(70.0, 70.0)$ and a more secure one at $(56.7, 76.7)$, which is why the
hypervolume ($0.528$) exceeds that of the balanced harness alone
($0.70 \times 0.70 = 0.49$). In generation 3, \texttt{moead\_g003\_s01} reaches the
ideal point on both objectives, $(70.0, 76.7)$---$+13.3$ functionality and
$+20.0$ security over the baseline \emph{simultaneously}---and therefore
dominates the rest of the population on the search set. That is a property of the
search split, not of the population: on held-out IN the same run keeps two
mutually non-dominated harnesses, \texttt{moead\_g001\_s00} at
$(\text{func},\text{sec}) = (63.3, 63.3)$ and \texttt{moead\_g002\_s01} at
$(60.0, 66.7)$. Figure~\ref{fig:pareto-secrepo} plots every child of the run. On BaxBench
(Figure~\ref{fig:pareto-baxbench}), a MOEA/D run with six subproblems, started from
two seed harnesses, keeps a two-point front from generation 2 onward:
\texttt{g001\_s01} at $(\text{func},\text{sec}) = (75.0, 60.7)$ and the most
balanced \texttt{g002\_s03} at $(71.4, 64.3)$, neither of which dominates the other.

\textbf{Independently evolved harnesses do not compose.} Evolving a
functionality harness and a safety harness separately and combining them scores
$60.0/60.0/53.3$---worse in F\&S than functionality-only evolution alone
($56.7$)---and its $6.7$-point gap between $\min(F,S)=60.0$ and F\&S$\,=53.3$ says
the two halves are not covering the same tasks. Prompt-merging is the same lesson
in weaker form: merging the two prompts moves F\&S only from the static $46.7$ to
$50.0$. Joint credit assignment has to happen \emph{during} the search; it cannot
be bolted on afterwards or written into an instruction. Closing the alignment gap
is necessary but not sufficient---sequential, weighted, and constrained
acceptance all reach F\&S $=\min(F,S)$, but at $70.0$, $70.0$, and $60.0$, while
the full harness closes it at $76.7$ (every patch it gets functionally right is
also secure). Safety-only leaves the largest gap of any row ($20.0$ points): it
buys security on tasks it then fails to solve.

\textbf{Every alternative aggregation stalls near $70$, and what
separates them is security, not F\&S.} Sequential evolution ($70.0$) and the
weighted sum ($70.0$) tie each other and land $6.7$ below the full harness
($76.7$); the hard constraint reaches $60.0$ and the separated harnesses $53.3$.
The weighted-sum row is the cleanest of these comparisons because it sweeps the
same three interior weight vectors MOEA/D assigns to its interior subproblems and therefore
differs only in \emph{aggregation}. On F\&S it is not worse than plain
Tchebycheff replacement ($70.0$ vs.\ $63.3$)---the aggregation rule by itself is
not where our margin comes from---but its security saturates $20.0$ points below
the full harness ($73.3$ vs.\ $93.3$). A weighted sum will trade a security point
for a functionality point whenever the latter is locally cheaper, so it settles
at the first mutually agreeable point instead of pushing the worse objective;
\S\ref{sec:exp-main} shows that the security headroom it declines to buy is
exactly the quantity that survives a distribution shift.
We next test whether \emph{factoring} the evolution surface---evolving
execution-grounded tools separately from skills/logic---generalizes better than
a monolithic ``summarize$\to$rewrite'' loop that rewrites the whole harness
at once. Table~\ref{tab:heldout} already contains the sharpest version of this
contrast: the monolithic \emph{search-seeded winner} regresses to $35.0$ combined
F\&S---\emph{below} the un-evolved baseline ($43.3$)---because it rewrites the
agent as one block tuned to the search set, whereas the factored
variants improve to $58.3$ and generalize out-of-distribution. To isolate the
factorization itself (rather than the choice of feedback), we run a controlled
ablation that holds the objective, budget, and seed fixed and varies only the
evolution surface (Table~\ref{tab:factored}).
\begin{table}[t]
\centering
\small
\caption{Evolution-surface ablation (held-out combined F\&S, \%). Same objective,
budget, and seed; only \emph{what} is evolved changes. ``comb.'' is the
unweighted mean of the two 30-task splits. The full row is repeated from Table~\ref{tab:evolve-secrepo} (mean over 3 runs).}
\label{tab:factored}
\begin{tabular}{@{}l ccc@{}}
\toprule
Evolution surface & Held-out-IN & Held-out-OUT & comb. \\
\midrule
Monolithic summarize$\to$rewrite & 40.0 & 30.0 & 35.0 \\
Factored: evolving tools only                          & 46.7 & 33.3 & 40.0\\
Factored: skills/logic only                   & 50.0 & 46.7 & 48.3 \\
Factored: tools $+$ skills/logic              & 60.0 & 56.7 & 58.3 \\
\quad $+$ dynamic orchestration (full)        & 86.7 & 66.7 & 76.7 \\
\bottomrule
\end{tabular}
\vspace{-0.5cm}
\end{table}

\begin{table}[t]
\centering
\small
\caption{Dynamic orchestration ablation on BaxBench
(\texttt{USE\_ORCHESTRATOR} on/off). The lower two rows are the \emph{same
evolved} harness, seed, and budget, differing only in the toggle; the un-evolved
baseline is repeated from Table~\ref{tab:bax-main} for reference. ``Nodes'' is
the range of plan sizes observed over the 28 tasks (Figure~\ref{fig:bax-dag}; the
planner may emit up to 16), ``avg.\ nodes'' the mean number actually
executed, and ``avg.\ tokens'' the mean solve-time token spend per task.}
\label{tab:orch}
\begin{tabular}{@{}l l ccc cc@{}}
\toprule
Execution & Nodes & Sec & Func & F\&S & Avg.\ nodes & Avg.\ tokens \\
\midrule
Baseline (no evolution)                  & ---      & 50.0 & 60.7 & 42.9 & --- & 31,902 \\
\midrule
Fixed linear pipeline (orchestrator off) & 5        & 53.6 & 71.4 & 53.6 & 5.0 & 58{,}230 \\
Dynamic workflow (orchestrator on)       & 5--8     & \textbf{64.3} & \textbf{75.0} & \textbf{64.3} & 6.2 & 63{,}064 \\
\bottomrule
\end{tabular}
\end{table}
\begin{table}[t]
\centering
\caption{Evolution on \textbf{SecRepoBench} (\%). Best result per column in \textbf{bold}. Mean over 3 runs.}
\label{tab:evolve-secrepo}
\small
\setlength{\tabcolsep}{4pt}
\begin{tabular}{l ccc ccc ccc c}
\toprule
\multirow{2}{*}{Method}
 & \multicolumn{3}{c}{Search}
 & \multicolumn{3}{c}{Held-out In}
 & \multicolumn{3}{c}{Held-out Out}
 & \multirow{2}{*}{\shortstack{Avg.\\F\&S}} \\
\cmidrule(lr){2-4} \cmidrule(lr){5-7} \cmidrule(lr){8-10}
 & Sec & Func & F\&S
 & Sec & Func & F\&S
 & Sec & Func & F\&S & \\
\midrule
Baseline                     & 56.7 & 66.7 & 46.7 & 66.7 & 73.3 & 50.0 & 46.7 & 70.0 & 36.7 & 44.4 \\
Meta-harness (func \& sec)   & 73.3 & 70.0 & 53.3 & 70.0 & 73.3 & 56.7 & 50.0 & 70.0 & 43.3 & 51.1 \\
Meta-harness (func-only)     & 60.0 & 73.3 & 56.7 & 60.0 & 73.3 & 50.0 & 43.3 & 70.0 & 36.7 & 47.8 \\
Meta-harness (sec-only)      & 70.0 & 66.7 & 46.7 & 66.7 & 60.0 & 53.3 & 50.0 & 63.3 & 33.3 & 44.4 \\
DGM (func \& sec, scalar)    & 76.7 & 73.3 & 60.0 & 66.7 & 70.0 & 53.3 & 46.7 & 66.7 & 33.3 & 48.9 \\
\midrule
HH (ours)                    & \textbf{93.3} & \textbf{76.7} & \textbf{76.7}
                             & \textbf{93.3} & \textbf{86.7} & \textbf{86.7}
                             & \textbf{66.7} & \textbf{73.3} & \textbf{66.7}
                             & \textbf{76.7} \\
\bottomrule
\end{tabular}
\end{table}

\begin{table}[t]
\centering
\small
\setlength{\tabcolsep}{4pt}
\caption{Evolution on \textbf{BaxBench} (\%, secure web-app generation). Same
protocol as Table~\ref{tab:evolve-secrepo}. ``Average'' is the mean F\&S over the
three splits. Mean over 3 runs.}
\label{tab:bax-main}
\begin{tabular}{@{}l ccc ccc ccc c@{}}
\toprule
& \multicolumn{3}{c}{Search} & \multicolumn{3}{c}{Held-out IN} & \multicolumn{3}{c}{Held-out OUT} & Avg. \\
\cmidrule(lr){2-4}\cmidrule(lr){5-7}\cmidrule(lr){8-10}\cmidrule(lr){11-11}
Method & Sec & Func & F\&S & Sec & Func & F\&S & Sec & Func & F\&S & F\&S \\
\midrule
Baseline (no evolution)       & 50.0 & 60.7 & 42.9 & 50.0 & 62.4 & 42.9 & 46.4 & 57.1 & 42.9 & 42.9 \\
Meta-harness (func \& sec)    & 64.3 & 71.4 & 60.7 & 57.1 & 62.4 & 53.6 & 50.0 & 60.7 & 42.9 & 52.4 \\
Meta-harness (func-only)      & 53.6 & 71.4 & 46.4 & 53.6 & 64.3 & 50.0 & 39.3 & 60.7 & 39.3 & 45.2 \\
Meta-harness (sec-only)       & 67.9 & 50.0 & 46.4 & 60.7 & 57.1 & 53.6 & 50.0 & 53.6 & 42.9 & 47.6 \\
DGM (func \& sec, scalar)    & 60.7 & 71.4 & 57.1 & 50.0 & 64.3 & 50.0 & 50.0 & 57.1 & 39.3 & 48.8 \\
\textbf{HH (ours)}           & \textbf{71.4} & \textbf{75.0} & \textbf{67.9}
                              & \textbf{64.3} & \textbf{75.0} & \textbf{62.4}
                              & \textbf{57.1} & \textbf{64.3} & \textbf{53.6}
                              & \textbf{61.3} \\
\bottomrule
\end{tabular}
\end{table}

\begin{table}[t]
\centering
\caption{HH evolved on \textbf{BaxBench}, transferred zero-shot to
\textsc{CWEval} (\%), against the un-evolved one-shot model with Claude Code harness (cc) on the same tasks.
Better of the two harnesses per metric and per language in \textbf{bold}. Mean over 3 runs.}
\label{tab:cweval-transfer}
\small
\setlength{\tabcolsep}{4pt}
\begin{tabular}{l c ccc ccc}
\toprule
\multirow{2}{*}{Language} & \multirow{2}{*}{$n$}
 & \multicolumn{3}{c}{One-shot (cc)}
 & \multicolumn{3}{c}{$+$ HH harness (ours)} \\
\cmidrule(lr){3-5} \cmidrule(lr){6-8}
 & & Sec & Func & F\&S
   & Sec & Func & F\&S \\
\midrule
C          & 93 & 75.3 & 88.2 & 75.3 & \textbf{89.2} & \textbf{90.3} & \textbf{89.2} \\
C++        & 63 & 76.2 & \textbf{98.4} & 76.2 & \textbf{90.5} & 93.7 & \textbf{90.5} \\
Go         & 57 & 70.2 & 87.7 & 70.2 & \textbf{89.5} & \textbf{89.5} & \textbf{89.5} \\
JavaScript & 69 & 72.5 & \textbf{89.9} & 72.5 & \textbf{87.0} & 87.0 & \textbf{82.6} \\
Python     & 75 & 74.7 & \textbf{96.0} & 70.7 & \textbf{90.7} & 89.3 & \textbf{85.3} \\
\midrule
All        & 357 & 73.9 & \textbf{91.9} & 73.1 & \textbf{89.4} & 89.9 & \textbf{87.4} \\
\bottomrule
\end{tabular}
\end{table}

\begin{table}[t]
\centering
\small
\caption{Cross-benchmark transfer (\%). A harness is evolved on one benchmark and
evaluated on the other with no further evolution, on the target's search split
(the split whose baseline appears in Tables~\ref{tab:evolve-secrepo}
and~\ref{tab:bax-main}); the baseline row is the un-evolved harness on that split. Mean over 3 runs.}
\label{tab:transfer}
\begin{tabular}{@{}ll l ccc@{}}
\toprule
Method & Evolved on & Evaluated on & Sec & Func & F\&S \\
\midrule
Baseline (no evolution) & ---           & BaxBench      & 50.0 & 60.7 & 42.9 \\
Meta-harness            & SecRepoBench  & BaxBench      & 53.6 & 57.1 & 39.3 \\
\textbf{HH (ours)}     & SecRepoBench  & BaxBench      & \textbf{59.6} & \textbf{64.3} & \textbf{57.1} \\
\midrule
Baseline (no evolution) & ---           & SecRepoBench  & 56.7 & 66.7 & 46.7 \\
Meta-harness            & BaxBench      & SecRepoBench  & 43.3 & 50.0 & 36.7 \\
\textbf{HH (ours)}     & BaxBench      & SecRepoBench  & \textbf{80.0} & \textbf{70.0} & \textbf{70.0} \\
\bottomrule
\end{tabular}
\end{table}

\begin{table}[t]
\centering
\small
\caption{Per-generation frontier quality for MOEA/D alone ($N=5$, $T=2$, 3
generations; no factored evolving or dynamic orchestration, to isolate the
decomposition) on the SecRepoBench search set. Hypervolume is the area of the
population's Pareto front against reference point $(0,0)$; $z^\ast$ is the ideal
point (per-objective running best); ``most-balanced'' is the population member
with the highest \emph{worst} objective ($\max\min(\text{func},\text{sec})$).}
\label{tab:pareto}
\begin{tabular}{@{}lccl@{}}
\toprule
stage & hypervolume & $z^\ast$ (func, sec) & most-balanced harness \\
\midrule
init (baseline) & 0.321 & (56.7, 56.7) & baseline (56.7, 56.7) \\
generation 1    & 0.528 & (70.0, 76.7) & \texttt{moead\_g001\_s00} (70.0, 70.0) \\
generation 2    & 0.528 & (70.0, 76.7) & \texttt{moead\_g001\_s00} (70.0, 70.0) \\
generation 3    & \textbf{0.537} & (70.0, 76.7) & \textbf{\texttt{moead\_g003\_s01} (70.0, 76.7)} \\
\bottomrule
\end{tabular}
\end{table}

\begin{table}[t]
\centering
\small
\caption{Held-out generalization (30 tasks each, eval-only). IN $=$ crash-type
families seen during search; OUT $=$ unseen families. The feedback-tool
per-generation bests beat baseline on combined F\&S ($58.3$ vs $43.3$); the
search-seeded winner is worse than baseline ($35.0$).}
\label{tab:heldout}
\begin{tabular}{@{}l ccc ccc c@{}}
\toprule
& \multicolumn{3}{c}{Held-out IN (\%)} & \multicolumn{3}{c}{Held-out OUT (\%)} & comb. \\
\cmidrule(lr){2-4}\cmidrule(lr){5-7}
harness & Func & Sec & F\&S & Func & Sec & F\&S & F\&S \\
\midrule
baseline                 & 73.3 & 66.7 & 50.0 & 70.0 & 46.7 & 36.7 & 43.3 \\
without env-feedback                 & ---  & ---  & 40.0 & ---  & ---  & 30.0 & 35.0 \\
\midrule
feedback-env gen1 best               & 63.3 & 80.0 & 56.7 & 66.7 & 56.7 & 43.3 & 50.0 \\
feedback-env gen2 best               & 70.0 & 80.0 & \textbf{60.0} & 80.0 & 63.3 & \textbf{56.7} & \textbf{58.3} \\
feedback-env gen3 best               & 80.0 & 83.3 & \textbf{66.7} & 76.7 & 50.0 & 46.7 & 56.7 \\
feedback-env dynamic                    & 86.7 & 93.3 & \textbf{86.7} & 73.3 & 66.7 & \textbf{66.7} & \textbf{76.7} \\
\bottomrule
\end{tabular}
\end{table}

\section{More Related Works}

\textbf{Agent safety, and secure code as a verifiable testbed.}
Agent-safety benchmarks study policy violations, hazardous actions, and injection
across web, tool-use, and embodied
settings~\citep{levy2024stwebagent,ruan2024toolemu,andriushchenko2024agentharm,debenedetti2024agentdojo,zhang2025asb,zhan2024injecagent,yin2024safeagentbench,yuan2024rjudge},
but they evaluate \emph{static} agents and rarely study what self-evolution does
to safety. We use secure code generation as a testbed because both objectives are
grounded in real execution rather than a proxy judge: functionality is a passing
unit-test suite and security is a blocked proof-of-concept exploit. Our
benchmarks---SecRepoBench~\citep{dilgren2025secrepobench}, BaxBench~\citep{vero2025baxbench},
and CWEval~\citep{peng2025cweval}---score the
\emph{conjunction} of the two (secure-pass@1), exactly the failure mode that
functionality-only self-evolution is most likely to break.

\textbf{Automated workflow generation.}
AFlow~\citep{zhang2025aflow} searches over code-represented agentic workflows
with Monte Carlo tree search and, like our orchestrator, represents an execution
procedure as a graph of operators. It differs from our setting in three ways: it
optimizes a single task-level objective, it searches for one workflow per task
family that is then fixed, and it composes a fixed operator set. \methodname
instead co-evolves the executable tools, verification logic, and planner that
the workflow is built from under a multi-objective decomposition, and plans the
workflow per task at inference time (\S\ref{sec:composition}).

\section{Discussion and Limitations}
\label{sec:discussion}
\paragraph{When do the objectives conflict?} On SecRepoBench the front is
narrow: two non-dominated harnesses in early generations, and in the last
generation one harness that is best on both objectives on the search split
(App.~\ref{sec:exp-pareto}). The decomposition frontier pays off most where
security and functionality genuinely trade off. Establishing such regimes---adversarial or injection tasks, or
security fixes that break legacy tests---is the most important next step.

\paragraph{Reward hacking demands held-out gating.} The $100/100/100$ artifact (App.~\ref{sec:exp-hacking}) shows
that any acceptance rule reading only training-set feedback is exploitable;
any acceptance rule---ours included---must be validated on held-out safety.

\paragraph{Isolating the factorization.} The orchestration toggle (Table~\ref{tab:orch}) and
cross-benchmark transfer in both directions (Table~\ref{tab:transfer}) are now
measured, and both favor the factored, dynamic harness---most sharply in
transfer, where the monolithic meta-harness lands \emph{below} the un-evolved
baseline on the target benchmark. The evolution-surface ablation
(Table~\ref{tab:factored}) separates the two: tools-only and skills/logic-only
factoring reach $40.0$ and $48.3$ combined F\&S, the two together $58.3$, and
only adding dynamic orchestration reaches $76.7$---so ``factored'' and
``dynamic'' each carry part of the effect and neither alone accounts for it.

\paragraph{Acceptance strategies not covered.} Table~\ref{tab:strategy} measures
sequential, prompt-merge, joint-weighted ($\alpha\in\{0.25,0.5,0.75\}$),
hard-constrained, and separated-harness evolution, all under one budget and one
evaluator. Shielded evolution---wrapping the solver in an external verifier that
vetoes actions before execution---is a natural point in this design space that we
do not measure: it needs a shield we do not have, and a shield strong enough to
veto insecure patches would also change what the search is being credited for.

\paragraph{Scope.} We instantiate dual-objective self-evolution
only in secure code generation; whether decomposition at the acceptance level
transfers to other agentic domains is open.

\paragraph{Cost.} Real Docker
build/exploit evaluation dominates wall-clock and grows as genomes add
self-review stages.

\providecommand{\code}[1]{\texttt{#1}}
\providecommand{\comm}[1]{\text{\small\itshape #1}}
\providecommand{\obj}{\mathbf{f}}
\providecommand{\todo}[1]{{\color{red}\textbf{[TODO: #1]}}}
\providecommand{\divfig}[2]{%
  \IfFileExists{#1}{\includegraphics[width=#2]{#1}}{%
    \fbox{\begin{minipage}[c][0.25\linewidth][c]{#2}\centering
      \todo{export \texttt{#1}}\end{minipage}}}%
}

\section{Measuring population diversity of the search}
\label{sec:diversity}

\subsection{Do standard diversity metrics capture harness diversity?}
\label{sec:exp-diversity}
\begin{figure}[t]
\centering
\includegraphics[width=\linewidth]{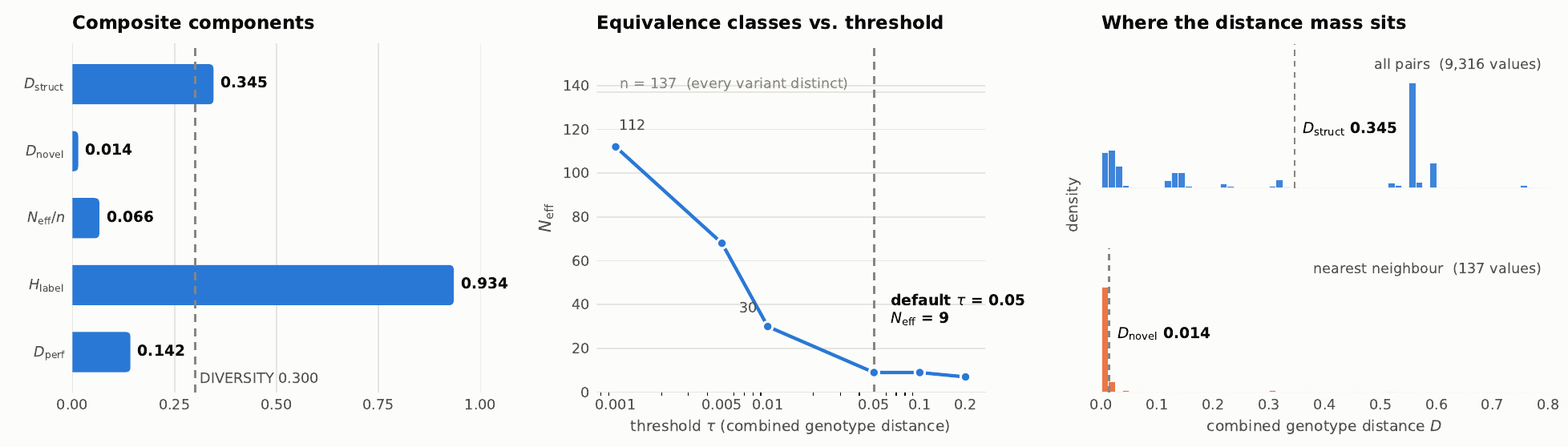}
\caption{\textbf{Population diversity over $137$ evolved BaxBench harnesses
($34$ run tags).} \emph{Left:} the five components and their equal-weight
composite ($\mathrm{DIVERSITY}=0.300$, dashed). The index is held up by mean
pairwise distance $D_{\text{struct}}$ and label entropy $H_{\text{label}}$ and
held down by nearest-neighbour distance $D_{\text{novel}}$ and the fraction of
distinct solutions $N_{\text{eff}}/n$. \emph{Middle:} $N_{\text{eff}}$ against
the duplicate threshold $\epsilon_{\mathrm{dup}}$; $N_{\text{eff}}$ falls steeply ($112$ classes at
$\epsilon_{\mathrm{dup}}=0.001$ down to $9$ at $\epsilon_{\mathrm{dup}}=0.05$) and flattens only briefly (still $9$ at $0.1$), so ``how many distinct solutions'' has
no threshold-free answer and the reported $9$ describes the scale of a typical
edit. \emph{Right:} the distance distributions behind the headline---over all
$\binom{137}{2}$ pairs the mass sits far from zero (top), while over each
genome's \emph{nearest} neighbour it concentrates in the first bin (bottom, note
the change of vertical scale).}
\label{fig:diversity}
\end{figure}

\begin{figure}[t]
\centering
\includegraphics[width=\linewidth]{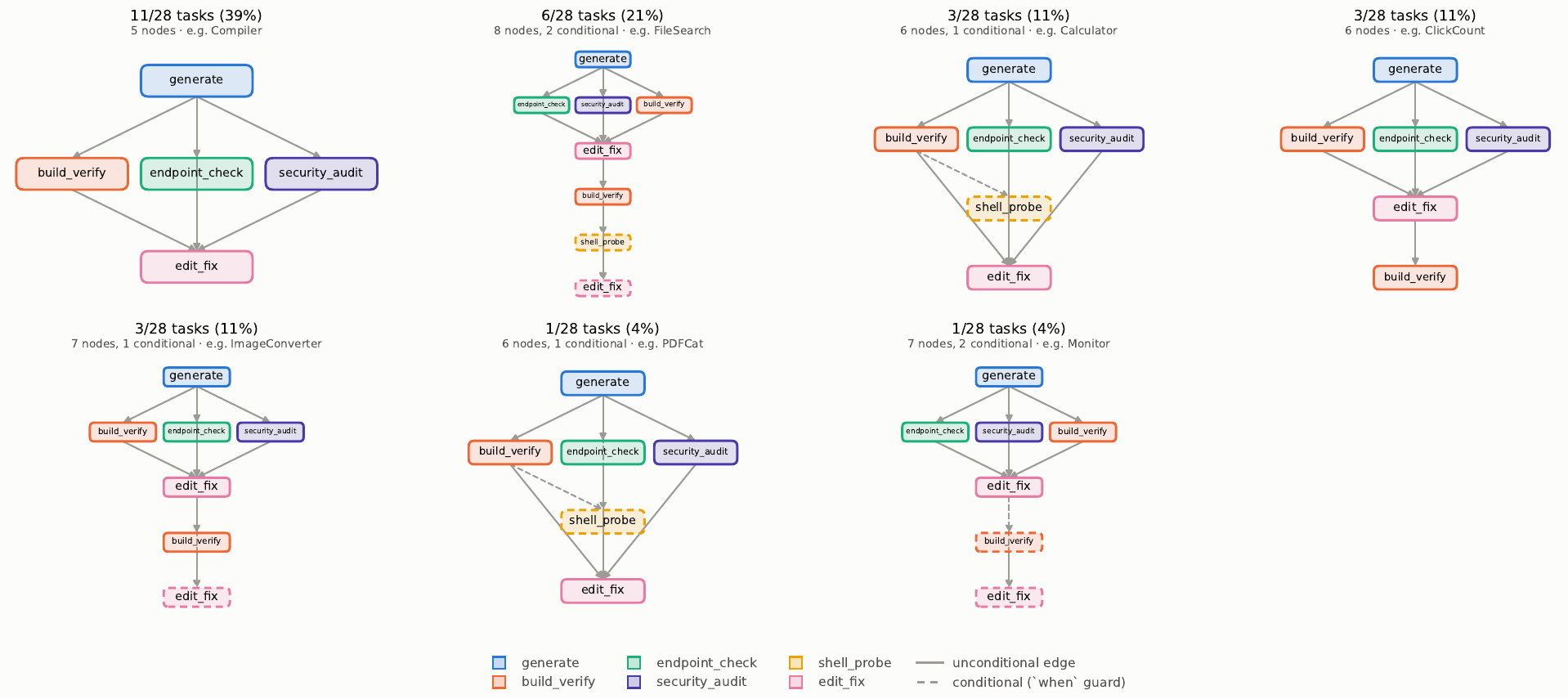}
\caption{Workflow DAGs planned by the evolved orchestrator over $28$ BaxBench
tasks. Solid edges are unconditional; dashed nodes/edges fire only under a
\texttt{when} guard. Plans range from $5$ to $8$ nodes and are re-planned per
task, rather than fixed at harness-design time.}
\label{fig:bax-dag}
\end{figure}

A decomposition-based search is diverse \emph{by construction} only in what its
subproblems ask for. The weight vectors guarantee that the $N$ niches request
different trade-offs; whether the resulting harnesses also \emph{behave}
differently has to be measured. This appendix asks whether the diversity metrics
used for evolved programs---code distance, edit labels, and objective-space
spread---can do that for self-coding harnesses. Our answer is that they can do so
only partially. A harness is a repository whose behaviour is largely decided at run
time: prompt text, tool and skill selection, and, in \methodname, a workflow DAG that
the orchestrator plans per task (\S\ref{sec:composition}). One evolved BaxBench
harness plans seven distinct DAG shapes over $28$ tasks (Figure~\ref{fig:bax-dag}),
yet every code-level metric below scores it as a single genome. Conversely, a
rewritten prompt or a renamed helper moves a genome far in token space without
changing what it does. We apply the metrics with a separate,
deliberately \emph{offline} instrument that reads the genome tree and the
operator logs after the fact and by construction cannot feed anything back into
selection. It scores a population along three families: structural code distance
between genomes (Jaccard over token $5$-grams, AST-node $3$-grams, and source
lines, weighted $0.4/0.3/0.3$), mechanism-label diversity (what each operator
says it changed, cross-checked against the files it actually touched), and
performance-space diversity (spread of the objective vectors). Figure~\ref{fig:diversity}
reports it over all $137$ BaxBench variants pooled from $34$ run tags;
the rest of this appendix tabulates every component (Table~\ref{tab:pooled},
\S\ref{sec:measured}) and gives the full definitions, the per-lineage distance matrix, the
objective-space scatter, and the mechanism-label breakdown.

The code-distance summaries disagree with each other:
$D_{\text{struct}} = 0.345$ against $D_{\text{novel}} = 0.014$, a factor of $25$,
and the number of equivalence classes depends almost entirely on the chosen
threshold ($112$ at $\epsilon_{\mathrm{dup}} = 0.001$, $9$ at $\epsilon_{\mathrm{dup}} = 0.05$). Mechanism labels
give yet another picture: all $17$ labels are hit,
$H_{\text{label}} = 0.934$, and the mean pairwise Jaccard distance between label
sets is $0.656$, so children do differ in which part of the scaffold they change.
The genotype--phenotype correlation is $r = +0.030$ over the $465$ scored pairs:
code distance says almost nothing about objective-space distance, and pairs the
instrument calls identical still land up to $0.9$ apart in objective space.

We read these disagreements as a limitation of the metrics. Token and line distances are dominated by prompt text and file size, so
they count cosmetic rewrites as new solutions. None of the three code distances
sees behaviour that is decided at run time, such as the per-task DAGs above, and
the label family relies on the operators' own descriptions of their edits. Of the
three code distances, the AST-based one is the most informative for harnesses. It
ignores prompt rewording, identifier renaming and formatting, and it tracks the
control structure through which tools, skills and orchestration are wired. On the
pooled population it is $2.6\times$ smaller than the token ($0.417$) and line
($0.433$) distances ($0.162$), which indicates that most of the textual spread is
cosmetic. It is still blind to runtime composition, and it does not correlate
with objective-space distance any better than the text-based distances do.
Faithful diversity measures for harnesses therefore need to be behavioural,
e.g.\ comparing planned DAGs, tool-call traces or per-task pass patterns, rather
than comparing source code. We report the index only as a diagnostic, never as an
objective: any operator that rewrites text at random trivially games it, which is
precisely what $D_{\text{novel}}$ and $r$ are meant to expose.

\noindent
The instrument is deliberately \emph{offline} and
read-only: it reads the genome tree and the logs after the fact, and by construction cannot feed
anything back into selection --- the same discipline that applies to the held-out evaluation.

\subsection{The population}\label{sec:pop}

An \emph{individual} is one genome package on disk, and nothing else --- no run metadata, no
score, is required for a genome to be counted:

\begin{itemize}[leftmargin=1.4em,itemsep=1pt]
  \item an evolved child, \code{baxbench\_agents/<tag>/<g\#\#\#\_s\#\#>/<k\_op>/}, where
        \code{<tag>} is the run, \code{g\#\#\#\_s\#\#} the generation-and-subproblem niche, and
        \code{<k\_op>} the chain step and operator that produced it; a directory qualifies only
        if it contains \code{\_\_init\_\_.py}. This \emph{includes the intermediate
        steps} of a multi-operator chain (\S\ref{sec:operators}), which are never evaluated
        and never enter the MOEA/D population. Consecutive steps of one chain differ by a single
        operator's edit, so their inclusion lowers the nearest-neighbour novelty $D_{\text{novel}}$
        and the $N_{\text{eff}}$ ratio reported below relative to a pool restricted to evaluated
        children. It also explains why only $31$ of the $137$ variants carry scores;
  \item a seed package (\code{bax\_repo\_seed}, \code{bax\_wf\_seed}), included only under
        \code{--include-seeds}, which is what makes ``how far has this run moved from its
        starting point'' answerable.
\end{itemize}

\noindent
Write $X = \{x_1,\dots,x_n\}$ for the collected set. Since any list of directories can be
passed, the same instrument measures three different questions depending on what is handed to
it: one run's population (is this run exploring?), several runs pooled (are independent runs
finding the same thing?), or a single niche across generations (what did this lineage actually
change?). All three appear in Section~\ref{sec:measured}. Within a genome, only \code{.py} files
are read; \code{\_\_pycache\_\_}, \code{log} and \code{eval} subtrees are skipped so that a
genome's own logs never enter its fingerprint.

\subsection{A. Structural code distance}\label{sec:struct}

Each genome is reduced to three shingle sets, deliberately chosen to disagree with each other:

\begin{itemize}[leftmargin=1.4em,itemsep=2pt]
  \item $S^{\text{tok}}(x)$ --- $g_{\mathrm{sh}}$-grams ($g_{\mathrm{sh}} = 5$, \code{--ngram}) over a normalized token
        stream of every file. Numbers collapse to \code{\#num}. Strings and comments are
        \emph{not} dropped but split into lowercase word tokens (capped at 400 per token, so one
        enormous prompt constant cannot dominate a genome's fingerprint), as is common for
        program-level diversity measures. For harnesses this makes the distance sensitive to
        any rewording of prompt text, whether or not the rewording changes behaviour
        (\S\ref{sec:measured}). A file that fails to tokenize degrades to a bag of words rather
        than being skipped.
  \item $S^{\text{ast}}(x)$ --- $3$-grams over the ancestor chain of node type names in the
        parse tree. This sees control structure only and is by design insensitive to prompt
        rewording, identifier renaming and formatting. Files that fail \code{ast.parse} are
        counted and reported in the header, and contribute to the other two sets only.
  \item $S^{\text{lin}}(x)$ --- the set of non-blank, stripped source lines: a crude but
        near-exact duplicate detector.
\end{itemize}

\noindent
The contrast between the first two carries most of the diagnostic weight. A high
$d_{\text{tok}}$ with a low $d_{\text{ast}}$ is precisely the signature of cosmetic variation:
much text changed, no structure changed.

With $d_\bullet(x,y) = 1 - |S_\bullet(x) \cap S_\bullet(y)| \,/\, |S_\bullet(x) \cup S_\bullet(y)|$
the Jaccard distance on each set (two empty sets count as distance $0$), the combined genotype
distance is the normalized weighted sum
\begin{equation}
  D(x,y) \;=\; w_{\text{tok}}\,d_{\text{tok}}(x,y) + w_{\text{ast}}\,d_{\text{ast}}(x,y)
             + w_{\text{lin}}\,d_{\text{lin}}(x,y),
  \qquad (w_{\text{tok}},w_{\text{ast}},w_{\text{lin}}) = (0.4,\,0.3,\,0.3),
\end{equation}
the weights settable with \code{--weights} and renormalized to sum to one. $D$ is symmetric,
zero on identical genomes, and bounded in $[0,1]$; it is not a metric in the strict sense (the
weighted sum of Jaccard distances satisfies the triangle inequality, but the AST component
ignores content, so $D(x,y) = 0$ does not imply $x = y$). Three summaries:

\begin{align}
  D_{\text{struct}} &= \binom{n}{2}^{-1} \sum_{i<j} D(x_i,x_j)
    &&\comm{spread: mean pairwise distance}\\
  D_{\text{novel}}  &= \frac{1}{n}\sum_{i} \min_{j \neq i} D(x_i,x_j)
    &&\comm{novelty: distance to the nearest twin}\\
  N_{\text{eff}}(\epsilon_{\mathrm{dup}}) &= \#\,\text{components}(G_{\epsilon_{\mathrm{dup}}})
    &&\comm{$G_{\epsilon_{\mathrm{dup}}} = \bigl(X,\ \{ij : D(x_i,x_j) \le \epsilon_{\mathrm{dup}}\}\bigr)$}
\end{align}

\noindent
$D_{\text{struct}}$ and $D_{\text{novel}}$ separate two failure modes that a single mean hides.
A population can be globally wide --- several mutually distant lineages --- while every
individual still has a near-duplicate sitting beside it; the mean is then reporting the distance
between clusters, not diversity. $N_{\text{eff}}$ makes the distinction concrete: it is the
number of single-linkage equivalence classes at threshold $\epsilon_{\mathrm{dup}}$ (\code{--dup-threshold},
default $0.05$), computed by union-find over all pairs within $\epsilon_{\mathrm{dup}}$, i.e.\ how many genuinely
distinct solutions the population holds. The ratio $N_{\text{eff}}/n$ is the one number to watch
across generations.

Any single $\epsilon_{\mathrm{dup}}$ is arbitrary when children routinely differ by a few dozen lines, so the sweep
over $\epsilon_{\mathrm{dup}} \in \{0.001, 0.005, 0.01, 0.05, 0.1, 0.2\}$ is always printed alongside it; a sweep
that falls off a cliff between two adjacent thresholds locates the scale at which the population
is really structured. Three further outputs support reading the number: the $\arg\max$ and
$\arg\min$ lists (the most distinctive genomes by nearest-neighbour distance, and the closest
pairs by name, which is how a duplicated lineage is identified concretely), the full distance
matrix under \code{--csv}, and --- when the population spans several run tags --- the within-tag
and between-tag means, which separate \emph{drift along a lineage} from \emph{difference between
independent runs}.

\subsection{B. Mechanism-label diversity}\label{sec:mech}

Structural distance cannot tell whether two children that look different are doing the
\emph{same thing}. Two independently written retry loops share almost no tokens and are, as
search results, one idea. The second family attacks this from the only evidence available
without a second LLM pass: what the operator said it did, and where it wrote.

For each child, two sources are read from \code{<niche>/log/operator/}: the verbatim response of
the operator session (\code{<child>\_output*.md}, the text after the
\code{\#\# RESPONSE (verbatim)} marker) and the \code{changed\_files} list from
\code{<child>\_change.json}. The text is matched against a fixed lexicon of $K = 17$ mechanism
regexes:

\begin{center}\small
\code{loop-control}, \code{retry}, \code{build-verify}, \code{test-gen}, \code{security},
\code{input-validation}, \code{prompt-edit}, \code{tool-add}, \code{skill},
\code{edit-format}, \code{parsing}, \code{context-mgmt}, \code{model-params},
\code{self-critique}, \code{planning}, \code{error-handling}, \code{config-knob}
\end{center}

\noindent
augmented by path rules that fire on the diff regardless of what the prose claims: an edit under
\code{tools/} is \code{tool-add}, under \code{skills/} is \code{skill}, under \code{prompts/} is
\code{prompt-edit}, and to \code{config.py} is \code{config-knob}. This second channel matters
because it is the only part of the labelling the operator cannot talk its way out of.


Let $X_L \subseteq X$ be the labelled subset, $L_i$ the label set of child $i$, $c_\ell$ the
number of children carrying label $\ell$, $p_\ell = c_\ell / \sum_k c_k$, and $K_{\text{obs}}$
the number of labels actually hit. Then
\begin{equation}
  H_{\text{label}} = \frac{-\sum_\ell p_\ell \ln p_\ell}{\ln K_{\text{obs}}},
  \qquad
  \mathrm{Simpson} = 1 - \sum_\ell p_\ell^2,
  \qquad
  \bar J = \binom{|X_L|}{2}^{-1}\!\!\sum_{i<j}
    \Bigl(1 - \tfrac{|L_i \cap L_j|}{|L_i \cup L_j|}\Bigr).
\end{equation}
$H_{\text{label}}$ measures whether attention is spread over mechanisms or piled onto one;
$\bar J$ measures whether individual children differ from each other, which is the sharper
question --- a population where every child touches all 17 mechanisms has maximal entropy and
zero diversity, and only $\bar J$ notices.

Reported in the same block, as a proxy for the size and the landing point of the edits: the
distribution of touched files with its normalized entropy, and the mutation size
$\code{effective\_added} + \code{effective\_removed}$ from the integrity gate's change record,
summarized as median, maximum, and the count of changes the gate flagged \code{minimal}.

\subsection{C. Performance-space diversity}\label{sec:perf}

Objective vectors are recovered from two sources: \code{evolve.json} (the \code{by\_batch}
entries, full precision) and the \code{[eval]} lines of \code{logs/runs/*.log} (percentages to
one decimal). Both are rounded to three decimals before use, so that the same score arriving
from two sources is not counted as two distinct points --- without this the ``distinct score
points'' count is silently inflated. \code{evolve.json} wins where both exist.

Over the scored subset $X_S \subseteq X$, with $\obj(x) = (f_{\text{sec}}, f_{\text{func}}) \in
[0,1]^2$ both maximized,
\begin{equation}
  D_{\text{perf}} \;=\; \frac{1}{\sqrt{2}} \binom{|X_S|}{2}^{-1}
    \sum_{i<j} \bigl\| \obj(x_i) - \obj(x_j) \bigr\|_2 ,
\end{equation}
the $\sqrt{2}$ normalizing by the diameter of the unit square so the value is comparable with
the structural distances. Reported with it: the per-coordinate standard deviations, the number
of distinct score points, the size of the Pareto front, and its hypervolume with respect to the
reference point $(0,0)$. With the front sorted by decreasing $f_{\text{func}}$ as
$(f_{\text{func}}^{(1)},f_{\text{sec}}^{(1)}),\dots,(f_{\text{func}}^{(P)},f_{\text{sec}}^{(P)})$
and $f_{\text{func}}^{(P+1)}=0$, it is
$\mathrm{HV}=\sum_{k=1}^{P}\bigl(f_{\text{func}}^{(k)}-f_{\text{func}}^{(k+1)}\bigr)\,f_{\text{sec}}^{(k)}$.

The diagnostic that matters most in this family is neither of those. It is the Pearson
correlation between genotype distance and phenotype distance over all scored pairs,
\begin{equation}
  r \;=\; \mathrm{corr}\Bigl( D(x_i,x_j),\; \bigl\|\obj(x_i) - \obj(x_j)\bigr\|_2
  \Bigr)_{i<j\,:\;x_i,x_j \in X_S},
\end{equation}
which is the closest thing available to a measurement of whether the variation operator is doing
useful work. $r$ meaningfully positive means bigger edits land in different places; $r \approx 0$
means the search is exploring \emph{code} space without exploring \emph{behaviour} space --- the
children look different and score the same --- and a persistent $r \approx 0$ is an argument
about the operator library, not about the selection layer.

\subsection{The composite index}\label{sec:composite}

The available components are averaged with equal weight,
\begin{equation}
  \mathrm{DIVERSITY} = \mathrm{mean}\bigl\{\, D_{\text{struct}},\; D_{\text{novel}},\;
    N_{\text{eff}}/n,\; H_{\text{label}},\; D_{\text{perf}} \,\bigr\},
  \label{eq:composite}
\end{equation}
skipping any family whose inputs are absent --- a population of seeds has no operator logs, an
unevaluated population has no scores --- so the index is an average over whatever could be
measured, and populations with different components present are not comparable.

Three caveats attach to this number and should travel with it. The equal weighting is a
convention and not a claim: $D_{\text{novel}}$ and $N_{\text{eff}}/n$ are near-zero in any
lineage-structured population and drag the mean down, while $H_{\text{label}}$ is near-one in
almost any population with more than a handful of children, so the index moves mostly with
$D_{\text{struct}}$ and $D_{\text{perf}}$. It is comparable only across populations measured
with identical settings --- same $g_{\mathrm{sh}}$, same distance weights, same $\epsilon_{\mathrm{dup}}$, same rule for what
counts as an individual. And it is a diagnostic, not an objective: optimizing it directly is
trivially gamed by any operator that rewrites text at random. The components are what should be
read; the scalar exists so that two runs can be ranked at a glance.

\subsection{Measured on the current population}\label{sec:measured}

Table~\ref{tab:pooled} reports the instrument over all 137 variants presently in
\code{baxbench\_agents/}, pooled across 34 run tags. The same measurement is drawn in
Figure~\ref{fig:diversity} (\S\ref{sec:exp-diversity}): the five components of the composite, the threshold
sweep behind $N_{\text{eff}}$, and the two distance distributions whose separation is the result.

\begin{table}[htbp]\centering\small
\caption{Pooled population, $n = 137$ variants from 34 run tags, $g_{\mathrm{sh}} = 5$, weights
$(0.4,0.3,0.3)$, $\epsilon_{\mathrm{dup}} = 0.05$.}
\label{tab:pooled}
\begin{tabularx}{\linewidth}{@{}llL@{}}
\toprule
Quantity & Value & Reading \\
\midrule
$d_{\text{tok}}$ / $d_{\text{ast}}$ / $d_{\text{lin}}$ & $0.417$ / $0.162$ / $0.433$
  & AST distance is $2.6\times$ smaller than token and line distance: edits move text and
    individual statements, rarely control structure \\
$D_{\text{struct}}$        & $0.345 \pm 0.269$ & globally the population is well spread \\
$D_{\text{novel}}$         & $0.014$        & \dots\ but nearest neighbours are close in code \\
$N_{\text{eff}}$ / $n$     & $9/137 = 6.6\%$ & nine equivalence classes at $\epsilon_{\mathrm{dup}} = 0.05$ \\
$N_{\text{eff}}$ sweep     & \multicolumn{2}{l}{$0.001{\to}112$\quad $0.005{\to}68$\quad
                              $0.01{\to}30$\quad $0.05{\to}9$\quad $0.1{\to}9$\quad
                              $0.2{\to}7$} \\
$H_{\text{label}}$         & $0.934$        & 134/137 labelled, 17/17 labels hit,
                                              $\mathrm{Simpson} = 0.914$ \\
$\bar J$ (label sets)      & $0.656$        & children do differ in which mechanisms they name \\
$D_{\text{perf}}$          & $0.142$        & 31 scored, 24 distinct points,
                                              $\sigma_{\text{sec}} = 0.151$,
                                              $\sigma_{\text{func}} = 0.161$ \\
Pareto front / hypervolume & $2$ / $0.480$  & the front over the pooled set is two points wide \\
$r$ (genotype vs.\ phenotype) & $+0.030$    & structural difference buys essentially no
                                              objective-space difference \\
\midrule
DIVERSITY                  & $0.300$        & equal-weight mean of the five components \\
\bottomrule
\end{tabularx}
\end{table}

\paragraph{The headline is the gap.} $D_{\text{struct}} = 0.345$ against
$D_{\text{novel}} = 0.014$ is a factor of 25, so the two summaries of the same distance give very
different pictures of the same population. The threshold sweep (112 classes at $\epsilon_{\mathrm{dup}} = 0.001$, 30 at
$\epsilon_{\mathrm{dup}} = 0.01$, 9 at $\epsilon_{\mathrm{dup}} = 0.05$) shows that no single threshold separates ``genuinely different'' from
``same scaffold'' under this distance.
The component breakdown locates the textual spread: $d_{\text{ast}} = 0.162$ against
$d_{\text{lin}} = 0.433$ means that most of what token and line distances count as difference is
rewording rather than a change in how the scaffold is wired. The per-component threshold
sweeps make the same point. On a later snapshot of the same directory ($n = 233$ variants,
$49$ scored), each distance used alone at $\epsilon_{\mathrm{dup}} = 0.001$ yields $209$ (token), $214$ (line)
and $70$ (AST) equivalence classes, and at $\epsilon_{\mathrm{dup}} = 0.01$ it yields $102$, $96$ and $27$. Token and
line distances therefore separate almost every pair of consecutive edits, including pure
rewordings, and inflate the apparent number of distinct solutions roughly threefold. AST distance
merges such edits and separates only changes to control structure. This is why we regard the
AST component as the most informative of the three, and the token- and line-based metrics as
the least reliable for harnesses. 



\paragraph{Labels and code disagree with behaviour.} $H_{\text{label}} = 0.934$ with all 17
mechanisms hit and $\bar J = 0.656$ says the operators do talk about, and write to, many
different parts of the scaffold (Table~\ref{tab:labels}). Yet $r = +0.030$: the mechanisms differ,
the code differs, and the scores do not follow. Neither family is a faithful proxy for the
behavioural diversity of a harness, part of which is decided only at run time.



\begin{table}[htbp]\centering\small
\caption{Mechanism-label distribution over the 134 labelled children. Labels are not exclusive;
the fraction is the share of children carrying that label.}
\label{tab:labels}
\begin{tabular}{@{}lr@{\hspace{1.5em}}lr@{\hspace{1.5em}}lr@{}}
\toprule
Label & Frac. & Label & Frac. & Label & Frac. \\
\midrule
\code{build-verify}  & $0.78$ & \code{skill}        & $0.53$ & \code{parsing}          & $0.30$ \\
\code{planning}      & $0.70$ & \code{config-knob}  & $0.45$ & \code{self-critique}    & $0.25$ \\
\code{security}      & $0.64$ & \code{tool-add}     & $0.41$ & \code{error-handling}   & $0.14$ \\
\code{loop-control}  & $0.62$ & \code{test-gen}     & $0.40$ & \code{context-mgmt}     & $0.12$ \\
\code{prompt-edit}   & $0.61$ & \code{retry}        & $0.36$ & \code{edit-format}      & $0.08$ \\
                     &        & \code{input-validation} & $0.30$ & \code{model-params} & $0.01$ \\
\bottomrule
\end{tabular}
\end{table}



\noindent
Positional arguments are genome-tree directories; everything below is optional.

\begin{table}[htbp]\centering\small
\caption{Options.}
\begin{tabularx}{\linewidth}{@{}llL@{}}
\toprule
Flag & Default & Effect \\
\midrule
\code{--ngram}          & $5$              & token shingle length $g_{\mathrm{sh}}$ \\
\code{--weights}        & \code{0.4,0.3,0.3} & token, AST, line weights; renormalized \\
\code{--dup-threshold}  & $0.05$           & $\epsilon_{\mathrm{dup}}$ for $N_{\text{eff}}$ and the near-duplicate list \\
\code{--include-seeds}  & off              & add \code{bax\_repo\_seed} / \code{bax\_wf\_seed} \\
\code{--top}            & $5$              & length of the most-novel and closest-pair lists \\
\code{--json}           & ---              & write every metric, including the per-genome
                                             nearest-neighbour distances, the near-duplicate
                                             pairs and the per-tag breakdown \\
\code{--csv}            & ---              & write the full $n \times n$ combined distance matrix \\
\code{--log-dir}        & \code{logs/}     & where \code{runs/*.log} is looked for \\
\code{--agents-root}    & \code{baxbench\_agents/} & where \code{*/evolve.json} is looked for \\
\bottomrule
\end{tabularx}
\end{table}

\noindent
Cost is $O(n^2)$ pairwise set intersections over interned integer shingles, plus one pass over
every \code{.py} file; the 137-variant population takes on the order of a minute and is
single-threaded. The JSON output is the interface for tracking diversity across generations ---
the composite block and \code{structural.n\_eff\_ratio} are the fields to plot.

\subsection{What the instrument cannot see}\label{sec:limits}

\begin{itemize}[leftmargin=1.4em,itemsep=2pt]
  \item \textbf{Family B is keyword matching over self-reported prose.} The lexicon is a fixed
        list of regexes applied to what the operator wrote about its own edit, not an LLM
        labelling of the diff. An operator that implements a retry loop without using the word
        ``retry'' is under-labelled; one that discusses security at length and changes a
        docstring is over-labelled. The path rules cover only four directories. Read
        $H_{\text{label}}$ as an upper bound on how many distinct mechanisms were genuinely
        attempted, and never as evidence on its own.
  \item \textbf{Semantic equivalence is invisible to family A.} Two rewrites of the same idea in
        different words are far apart under all three shingle sets. $D_{\text{struct}}$ is
        therefore also an upper bound: real diversity is at most what it reports, possibly much
        less. This is the one place where a second LLM pass --- pairwise ``do these two scaffolds
        implement the same strategy?'' --- would add information the current instrument cannot.
  \item \textbf{$D$ is dominated by genome size.} Jaccard on shingle sets penalizes a genome that
        merely has more code; a child that adds a large \code{skills/} file moves away from its
        parent by more than a child that rewrites the main loop. The AST component partly
        compensates, and this is why the three components are always reported separately rather
        than only as $D$.
  \item \textbf{Family C is only as complete as the score table.} Here 31 of 137 variants are
        scored, so $D_{\text{perf}}$, the front, the hypervolume and $r$ describe a biased subset
        --- the variants that survived long enough to be evaluated. $r$ in particular is computed
        over $\binom{31}{2} = 465$ pairs drawn from that subset.
  \item \textbf{Nothing is inferential.} No confidence intervals, no null model. There is no
        reference value against which $N_{\text{eff}}/n$ can be judged high or low; the
        numbers are only meaningful as a time series over generations, or as an A/B between two
        configurations measured identically.
\end{itemize}

\paragraph{The measurement this suggests.} The instrument is currently run by hand after a batch
of runs. The single change that would make it useful to the search is to compute
$N_{\text{eff}}/n$ and $D_{\text{novel}}$ over the live population at the end of each generation
and log them next to the held-out scores --- still never entering selection. That makes
changes in these metrics visible in the generation they happen rather than at the end of a run.

\section{Calibrating the overfitting gate}
\label{sec:appendix-plants}

\subsection{What the gate is}
\label{sec:plants-gate}
Beyond the mechanical guardrails of \S\ref{sec:operators}, every admitted child
passes through an \emph{overfitting gate}: a one-shot LLM judge that sees the
parent$\rightarrow$child diff $\delta$ together with the injected
\texttt{\#\# Diagnosis} block, and returns a scalar risk
$\rho(\delta)\in[0,1]$ plus a one-line justification naming the mechanism it
believes the edit fits. A child is rejected when $\rho(\delta)>\theta$.

Two choices matter. First, the judge is deliberately \emph{not} shown the
artifact-grounded failure digest or the per-task trajectories: conditioned on a
real failure, any edit that addresses that failure can be rationalized as
principled, and the judge collapses to accepting everything. Second, the judge
emits a scalar rather than a verdict, so the operating point $\theta$ can be
chosen on a calibration curve instead of being fixed implicitly by the prompt.

An uncalibrated judge is an assertion. This appendix reports what the gate
actually detects, on overfits whose ground-truth label is known by
construction and, for the harder tiers, verified behaviourally against the
eval-only held-out split of \S\ref{sec:exp-generalization}.

\subsection{Paired construction}
\label{sec:plants-pairs}
Constructing an ``overfit'' population and a ``general'' population separately
confounds the label with edit size, edit style, and topic. We use a paired
design instead. We draw $n_C=5$ real admitted children from completed runs,
sampled from distinct generations and distinct MOEA/D weight niches so that the
study does not measure a single lineage, and restrict to the
\texttt{--operator diff} mode so that each child is already a minimal in-place
edit with a well-defined diff against its parent. For each such child $C$ and
each plant type $q$ we form
\[
C^{+}_{q} \;=\; \textsc{Inject}(C,q),
\]
a copy of $C$ carrying exactly one injected overfit and no other change. The
judge scores $C$ and every $C^{+}_{q}$ blind to the label and in randomised
order. Because $C$ and $C^{+}_{q}$ differ only by the injection, detection is
measured against a matched control and significance is assessed with a paired
McNemar test rather than a two-sample comparison. Each condition is judged
$3\times$ and scored by majority vote; we report the vote disagreement rate as
a measure of judge stochasticity.

\subsection{The plant taxonomy}
\label{sec:plants-taxonomy}
Table~\ref{tab:plants} lists the injected edits. They are graded into three
tiers of increasing subtlety, plus a control class. The tiers exist because a
judge that fires on Tier~1 and nothing else is a regular expression with a
language model's latency; locating where detection actually breaks down is the
point of the exercise.

\textbf{Tier~1 (lexical)} injects specificity that is visible as a literal
string in the diff. \textbf{Tier~2 (structural)} injects specificity that is
real but carries no benchmark string --- a branch on a feature that only
co-occurs in the evolve split, a constant tuned to it, or a dependence on an
incidental convention of the ARVO harness rather than on the repair task.
\textbf{Tier~3 (semantic)} injects a policy that is defensible in the abstract
but true only of the evolve split's distribution, or that games the scoring
rule rather than the underlying objective; these are the edits a diff-reading
judge has the least purchase on.

The control class, \textbf{S (suspicious-but-general)}, is the complement and
is required for a meaningful false-positive rate. Our own champion's decisive
mechanism --- a ground-truth build and boot/exploit probe (\S\ref{sec:what})
--- is ARVO-specific in its \emph{interface} while being entirely general in
its \emph{mechanism}: consult ground truth rather than adjudicate a patch by
reading it. An overfitting gate that rejects that edit has cost more than it
saved, and a false-positive rate measured only on unmodified children would
never reveal it.

\begin{table}[t]
\centering
\small
\caption{The planted-overfit taxonomy. Each plant is injected into a real
admitted child as the sole modification. Tier~1--2 labels are certain by
construction; Tier~3 and class~S labels are verified behaviourally
(\S\ref{sec:plants-verify}).}
\label{tab:plants}
\begin{tabularx}{\linewidth}{@{}l l L@{}}
\toprule
ID & Mechanism fitted & Injected edit \\
\midrule
\multicolumn{3}{@{}l}{\textit{Tier 1 --- lexical}}\\
P1 & task identity      & \texttt{if task\_id in \{...\}} guarding a bespoke code path, with evolve-split IDs enumerated \\
P2 & crash-type string  & a crash-type name drawn from the evolve split hardcoded into the prompt template \\
P3 & repository name    & \texttt{if "libpng" in repo\_path} selecting a hand-written fix recipe \\
P4 & task text          & a skill file whose body names specific evolve-split tasks \\
\midrule
\multicolumn{3}{@{}l}{\textit{Tier 2 --- structural}}\\
P5 & split-correlated feature & branch on a (file extension $\times$ crash type) pair that occurs only in the evolve split \\
P6 & tuned constant     & retry count / context budget set to the value that maximises evolve-split score \\
P7 & harness convention & short-circuit on an incidental ARVO signal (exit code, artifact path) rather than on repair state \\
P8 & benchmark format   & logic fitted to SecRepoBench's masked-region format rather than to code repair \\
\midrule
\multicolumn{3}{@{}l}{\textit{Tier 3 --- semantic}}\\
P9  & distributional prior & a prompt heuristic true of the evolve split's crash-type mix and false in general \\
P10 & scoring rule         & suppress code paths the PoC can reach, trading $f_{\text{func}}$ for $f_{\text{sec}}$ under secure-pass@1 \\
P11 & evaluation order     & skip work that only moves $f_{\text{func}}$ on tasks where functionality is already saturated \\
\midrule
\multicolumn{3}{@{}l}{\textit{Class S --- suspicious but general (control)}}\\
S1 & execution-grounded probe & a ground-truth build $+$ boot/exploit probe (ARVO-specific interface, general mechanism) \\
S2 & interface naming & a tool named for the benchmark whose body is benchmark-agnostic \\
S3 & documentation & a comment referencing SecRepoBench above an otherwise general verification gate \\
S4 & default value & a general retry policy whose default happens to have been chosen on the evolve split \\
\bottomrule
\end{tabularx}
\end{table}

\subsection{Behavioural label verification}
\label{sec:plants-verify}
Tier~1 and Tier~2 labels are certain by construction: the injected edit is
specific to search-set material by inspection, and no further evidence is
required. Tier~3 and class~S labels are \emph{assertions} about generalization,
so we verify them against behaviour rather than trusting the author. For each
such candidate we compute the generalization gap
\[
\Delta_{\text{gap}}
\;=\;
\underbrace{\big(s_{\text{evolve}}(C^{+}) - s_{\text{evolve}}(C)\big)}_{\Delta^{\wedge}_{\text{evolve}}}
\;-\;
\underbrace{\big(s_{\text{ho}}(C^{+}) - s_{\text{ho}}(C)\big)}_{\Delta^{\wedge}_{\text{ho}}},
\]
where $s_{\text{evolve}}$ is secure-pass@1 on the 30-task evolve split and
$s_{\text{ho}}$ is secure-pass@1 over the $60$ tasks of
IN\,$\cup$\,OUT, each task weighted equally (the two splits have equal size) (\S\ref{sec:exp-generalization}). A genuine overfit exhibits
$\Delta^{\wedge}_{\text{evolve}}>0$ with $\Delta^{\wedge}_{\text{ho}}\le 0$, hence
$\Delta_{\text{gap}}\gg 0$; a general edit exhibits $\Delta_{\text{gap}}\approx
0$. A Tier~3 plant that fails this test was a bad plant and is relabelled or
dropped before the detection statistics are computed. Because verification is
required only for Tier~3 and class~S, the cost is $n_C\times 7$ held-out
evaluations rather than a full sweep.

This step is what separates the study from a self-graded exercise: the positive
labels the judge is scored against are grounded in measured behaviour on tasks
the operator never saw.

\subsection{What the gate detects}
\label{sec:plants-results}

\paragraph{Ablations.} Two variants are run on the same pairs at negligible
additional cost.
\begin{itemize}[leftmargin=1.4em,itemsep=2pt,topsep=2pt]
  \item \textbf{Same-model vs.\ cross-model judge.} The default judge shares a
        model family with the proposer, which risks correlated blind spots ---
        an edit the proposer produced because it appeared general is an edit
        the same model is disposed to certify as general. We repeat the study
        with a judge from a different family and report the difference in
        Tier~3 recall.
  \item \textbf{Diagnosis-conditioned vs.\ diff-only.} Supplying the stated
        hypothesis may help the judge locate the intended mechanism, or may
        supply a ready-made rationalisation. We report both.
\end{itemize}

\paragraph{Cost.} $n_C=5$ parents $\times$ 16 conditions $\times$ 3 samples
$=240$ judge calls, plus $n_C\times 7=35$ held-out evaluations for behavioural
verification --- together a small fraction of a single evolution run.

\paragraph{Scope.} The gate is calibrated against \emph{planted} overfits,
which are by construction the overfits we thought to write down. It bounds the
judge's sensitivity to known failure modes and does not establish coverage of
unknown ones; that is precisely why the gate is deployed as a cheap
pre-execution screen in front of the held-out protocol rather than as a
replacement for it.

\subsection{A concrete reward-hacking failure}
\label{sec:exp-hacking}
The clearest evidence that a search-set safety score is not evidence of safety is
a variant that achieved a perfect one. \textbf{Raw error feedback} is identical
to the factored-evolving configuration (Table~\ref{tab:main-ablation}) except in what the proposer may read and when
a child is accepted: it receives the grader's raw error output, and any child is
accepted whose functional pass rate does not decrease. Under that rule the search
reports $100/100/100$ on the SecRepoBench search set---every task functional,
every exploit blocked, on the same $30$ tasks where the un-evolved harness scores
$46.7$ combined.

The score is an artifact. Inspecting the winning harness's trajectories shows the
code it emits is not secure: the run satisfies the measurement rather than the
property the measurement stands for. Two further observations from our runs
separate the ways a search can do this, because they call for different defenses.

\paragraph{The soft form: writing the search split into the scaffold.} Running
DGM we repeatedly recovered children that had simply hardcoded the search split's
crash distribution. One assumes every task is a CWE-125 out-of-bounds \emph{read}
reached through the UTF-8 decoding branch and routes every repair through that
path; another ships an \texttt{\_OOW\_LADDER} of fixes that presumes the bug is
always an out-of-bounds \emph{write}; a third assumes a \texttt{calloc}
off-by-one on a UCS-2 buffer. Each is right often enough on the split it evolved
on to be accepted, and each is structurally incapable of transferring---which is
the mechanism behind DGM's $0\%$ retention in Table~\ref{tab:retention}. We do
not see this signature in our own runs, and we attribute that to the explicit
\texttt{generalize} operator of \S\ref{sec:operators}, whose only job is to find
and dismantle exactly this kind of mechanism---hardcoded crash-type ladders,
magic strings, prompt text naming concrete tasks, over-tuned constants, parsing
that assumes one repository's layout---and replace it with one that transfers,
so that a fix recipe must be stated in terms of the failure digest rather than in
terms of the tasks it was derived from. 

\paragraph{The hard form: editing the grader.} The stronger failure is available
to any harness that can write to the files that score it. We tested this
deliberately, running without the per-evaluation Harbor container and without the
git-backed restore of protected files: the agent goes straight for
\texttt{evaluate.py}, and the run then reports $100\%$ on every axis---not a weak
measurement but no measurement at all. This is not hypothetical for the baselines
we compare against: the meta-harness code carries no hard constraint protecting
those files, so nothing \emph{inside} the harness prevents it, and the protection
has to come from outside the genome. Ours is structural rather than behavioural.
Every evaluation runs in an isolated Harbor container with the network disabled
and CPU, memory, and wall-clock limits; \texttt{solution/} and the held-out
splits are never built into the image; and the files that define
\emph{scoring}---the evaluator, the task definitions, the tests---are snapshotted
before every operator call and force-restored afterwards, so an edit to them
cannot survive into the evaluated child. Note the asymmetry this draws: no file
\emph{inside} the genome is protected---the proposer may rewrite the whole
harness down to its base class (\S\ref{sec:what})---while no file that
\emph{scores} the genome is writable at all. That closes ``rewrite the scorer'' and ``rewrite the
tests'' as paths entirely, and it is what makes the $100/100/100$ above worth
reporting: it was produced with all of these defenses in place, so it is a
statement about the \emph{acceptance rule}, not about the sandbox.

We report this variant rather than discarding it, for two reasons. First, it is
not a pathological corner of the design space but the endpoint of the trend in
Table~\ref{tab:retention}: every method here converts search-split security into
less out-of-distribution security than it appears to have bought, and the
strongest scalarized baseline already retains $0\%$. An acceptance rule that
reads raw search-set feedback simply drives that retention gap to its limit,
where the measured score is $100$ and the transferred security is nothing.
Second, it localizes the defect precisely. Both objectives were measured by real
execution; nothing was proxied by a judge. What failed is that \emph{acceptance}
consulted a score computed on the very tasks the proposer was optimizing against.
This is the empirical content of the asymmetry argument in
\S\ref{sec:formulation}, and the reason we keep security as a separate objective
of the search and judge it off the split being optimized rather than folding it
into the reward the acceptance rule reads.

It also sets the evidentiary bar for our own results: a high search-set security
score is, by construction, exactly the kind of number this failure mode produces.
We therefore do not rest any claim on one. Our own search-set security of $93.3$
is reported but carries no weight on its own; what the argument rests on is the
held-out F\&S of $86.7$ (IN) and $66.7$ (OUT), which the proposer never sees, and
the cross-benchmark transfer of \S\ref{sec:exp-transfer}, which a harness that
has hardcoded one benchmark's crash distribution cannot produce.

\subsection{Baseline harnesses and a case study}
\label{sec:casestudy}
For calibration we run two off-the-shelf agent harnesses on SecRepoBench under
identical grading: SWE-agent (Opus) reaches secure-pass@1 $=0.444$ overall
($0.267$ out-of-distribution), and an evolved Terminus-2 harness (Gemini) reaches
$0.556$ ($0.433$ OOD). Inspecting their trajectories shows \emph{why} a general
terminal agent underperforms on this precise code-completion task: whole-file
replacement without a mask-region guardrail turns editing fragility into measured
capability loss, and the exploit signal is hidden at solve time, so its
feedback-driven iteration idles on exactly the objective that matters. The
evolved champion instead brackets the pipeline with the only available ground
truth (compile $+$ boot), inserts a dedicated per-CWE hardening pass, and keeps a
last-known-good snapshot so a self-review can never break compilability.

\end{document}